\documentclass[10pt,twocolumn,letterpaper]{article}

\usepackage[pagenumbers]{cvpr} 

\usepackage{graphicx}
\usepackage{amsmath}
\usepackage{amssymb}
\usepackage{booktabs}
\usepackage{multirow}
\usepackage{enumitem}
\usepackage[super]{nth}

\usepackage{tcolorbox}
\definecolor{lightblue}{HTML}{18282e}
\definecolor{lighterblue}{HTML}{f2fafd}  
\newtcolorbox{abox}{colback=lighterblue,colframe=lightblue}

\usepackage{scalerel,xparse}

\usepackage[pagebackref,breaklinks,colorlinks]{hyperref}

\NewDocumentCommand\emojiturtle{}{
\raisebox{0.3\height}{\includegraphics[scale=1]{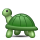}}
}
\NewDocumentCommand\emojiny{}{
\raisebox{0.35\height}{\includegraphics[trim={2cm 0 2cm 0},clip,scale=0.04]{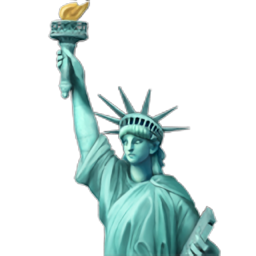}}
}

\usepackage[capitalize]{cleveref}
\crefname{section}{Sec.}{Secs.}
\Crefname{section}{Section}{Sections}
\Crefname{table}{Table}{Tables}
\crefname{table}{Tab.}{Tabs.}

\def\cvprPaperID{11309} 
\def\confName{CVPR}
\def\confYear{2023}

\begin{document}


\title{Diffusion Art or Digital Forgery? Investigating Data Replication\\ in Diffusion Models}

\author{Gowthami Somepalli\emojiturtle, 
Vasu Singla\emojiturtle, 
Micah Goldblum\emojiny,
Jonas Geiping\emojiturtle,
Tom Goldstein\emojiturtle
\\
\and 
\emojiturtle University of Maryland, College Park\\
{\tt\small \{gowthami, vsingla, jgeiping, tomg\}@cs.umd.edu}
\and
\emojiny New York University\\
{\tt\small goldblum@nyu.edu}
}


\twocolumn[{%
\renewcommand\twocolumn[1][]{#1}%
\maketitle
\begin{center}
    \centering
    
    \vspace{-1.0em}
    \includegraphics[width=\linewidth, trim = 3.5mm 2mm 0.0mm 4mm, clip]{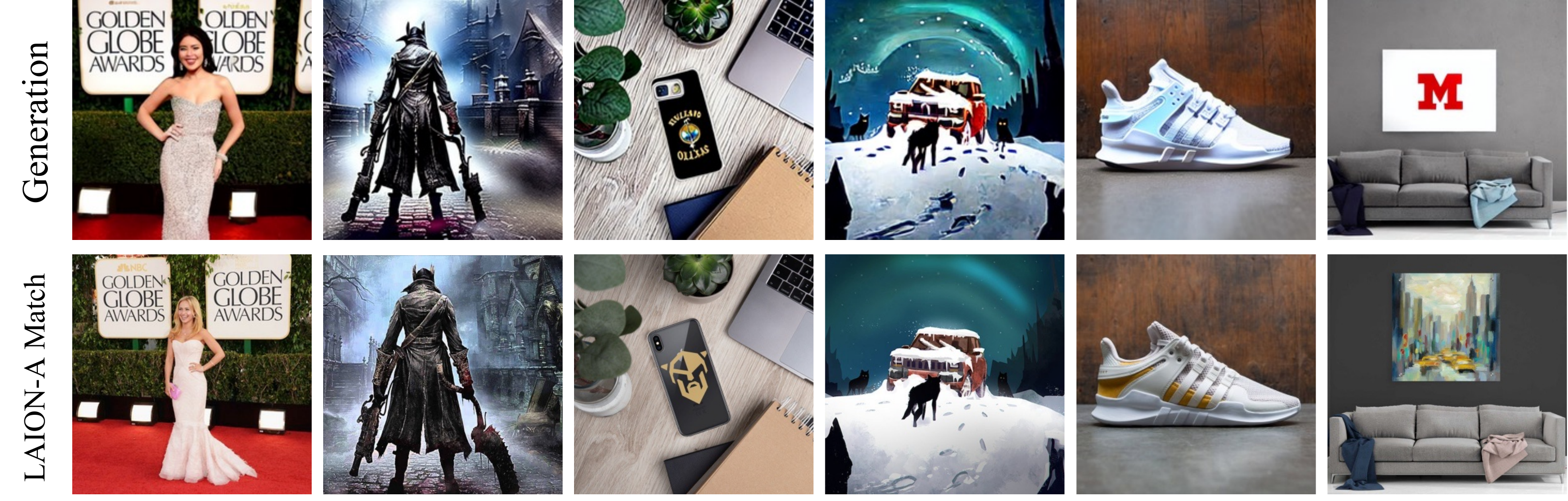} 
    \captionof{figure}{{\em Stable Diffusion} is capable of reproducing training data, creating images by piecing together foreground and background objects that it has memorized.  Furthermore, the system sometimes exhibits {\em reconstructive} memory, in which recalled objects are semantically equivalent to their source object without being pixel-wise identical. Here, we show this behavior occurring with a range of prompts sampled from LAION, and with a hand-crafted prompt (rightmost pair). The presence of such images raises questions about the nature of data memorization and the ownership of diffusion images. Top row: generated images. Bottom row: closest matches in the LAION-Aesthetics v2 $6+$ set. Sometimes source and match prompts are quite similar, and  sometimes they are quite different. See \Cref{fig:sd_train_prompts_copies} for more examples with prompts, or the Appendix for prompts from this figure.}
    \label{fig:teaser}
\vspace{0.5em}
\end{center}
}]

\begin{abstract}

Cutting-edge diffusion models produce images with high quality and customizability, enabling them to be used for commercial art and graphic design purposes. But do diffusion models create unique works of art, or are they replicating content directly from their training sets?  In this work, we study image retrieval frameworks that enable us to compare generated images with training samples and detect when content has been replicated.  Applying our frameworks to diffusion models trained on multiple datasets including Oxford flowers, Celeb-A, ImageNet, and LAION, we discuss how factors such as training set size impact rates of content replication.  We also identify cases where diffusion models, including the popular {\em Stable Diffusion} model, blatantly copy from their training data. 

\end{abstract}

\section{Introduction}

\label{sec:intro}

The rapid rise of diffusion models has led to new generative tools with the potential to be used for commercial art and graphic design. The power of the diffusion paradigm stems in large part from its reliance on simple denoising networks that maintain their stability when trained on huge web-scale datasets containing billions of image-caption pairs. 

These mega-datasets have the power to forge commercial models like \textit{DALL$\cdot$E} \cite{ramesh2022hierarchical} and {\em Stable Diffusion} \cite{rombach2022high}, but also bring with them a number of legal and ethical risks \cite{birhane_multimodal_2021}. Because these datasets are too large for careful human curation, the origins and intellectual property rights of the data sources are largely unknown.  This fact, combined with the ability of large models to memorize their training data \cite{feldman_what_2020,carlini_extracting_2021,carlini_quantifying_2022}, raises questions about the originality of diffusion outputs. There is a risk that diffusion models might, without notice, reproduce data from the training set directly, or present a collage of multiple training images. 

\looseness -1 We informally refer to the reproduction of training images, either in part or in whole, as \textit{content replication}. In principle, replicating partial or complete information from the training data has implications for the ethical and legal use of diffusion models in terms of attributions to artists and photographers. Replicants are either a benefit or a hazard; there may be situations where content replication is acceptable, desirable, or fair use, and others where it is ``stealing.'' While these ethical boundaries are unclear at this time, we focus on the scientific question of \textit{whether replication actually happens with modern state-of-the-art diffusion models, and to what degree}.

Our contributions are as follows. 
We begin with a study of how to detect content replication, and we consider a range of image similarity metrics developed in the self-supervised learning and image retrieval communities. We benchmark the performance of different image feature extractors using real and purpose-built synthetic datasets and show that state-of-the-art instance retrieval models work well for this task. 
 
Armed with new and existing tools, we search for data replication behavior in a range of diffusion models with different dataset properties. We show that for small and medium dataset sizes, replication happens frequently, while for a model trained on the large and diverse ImageNet dataset, replication seems undetectable.  

This latter finding may lead one to believe that replication is not a problem for large-scale models. However, the even larger {\em Stable Diffusion} model exhibits clear replication in various forms (Fig \ref{fig:teaser}).  Furthermore, we believe that the rate of content replication we identify in {\em Stable Diffusion} likely underestimates the true rate because the model is trained on a large 2B image split of LAION, but we only search for matches in the much smaller 12M ``Aesthetics v2 6+'' subset.

The level of image similarity required for something to count as ``replication'' is subjective and may depend on both the amount of diversity within the image's class as well as the observer. Some replication behaviors we uncover are unambiguous, while in other instances they fall into a gray area. Rather than choosing an arbitrary definition, we focus on presenting quantitative and qualitative results to the reader, leaving each person to draw their own conclusions based on their role and stake in the process of generative AI.

\section{Background}
\label{sec:background}
Below, we review the background and related work in image retrieval, generative models, and memorization literature.

\smallskip
\textbf{Image retrieval and copy detection.} The process of searching a database for images containing reference features from a source image is known as {\em image retrieval}. The related task of {\em inexact copy detection} requires high semantic similarity between the source and match \cite{douze2009evaluation}.  
Image retrieval works with image descriptors based on all types of neural networks \cite{babenko2014neural,razavian2016visual}. High-performance descriptors can be fine-tuned specifically for retrieval after unsupervised training  \cite{radenovic2018fine, radenovic2016cnn} using structure-from-motion (SfM) or contrastive objectives \cite{gordo2016deep,chen2022deep}. 
A natural basis for image retrieval methods are self-supervised models that inherently learn strong feature descriptors, matching similar images to similar representations \cite{chen2021empirical,chen2020simple,caron2018deep,grill2020bootstrap,he2022masked}. A particularly relevant SSL method for our purposes is DINO ~\cite{caron2021emerging}, which is shown to perform competitively on instance retrieval tasks.

Recent approaches adopt strong vision transformers as architectural backbones for retrieval~\cite{gong2014multi,el2021training,song2022boosting,berman2019multigrain,jang2021self}. Historical progress in this field is tracked by public image similarity challenges \cite{douze20212021}. A recent SOTA approach is SSCD~\cite{pizzi2022self}, which builds on previous work in self-supervised representation learning and optimizes a descriptor for copy detection using entropic regularization and an array of task-specific data augmentations.

\smallskip
\textbf{Memorization in deep learning.} While it is widely known and discussed that large models can memorize their data, there is no universally accepted definition of memorization. To ML theorists, memorization is synonymous with overfitting  \cite{arpit2017closer,feldman2020neural,feldman2020does}. 
In the field of membership inference attacks, one seeks to determine whether a chosen image was part of the training set 
\cite{hu_membership_2021,carlini_membership_2022,wen_canary_2022,webster_this_2021}.  Indeed, it has been shown that models retain a memory of the contents of their training set, particularly when training samples are repeated \cite{wen_canary_2022}. 
 Note that membership inference can be done by reconstructing original training data from the model \cite{webster_this_2021}, although this is not the goal of most membership inference methods.  The problem of explicitly reconstructing images from the training set of a classifier is known as {\em model inversion}, and recent research has been able to do this with both convolutional and transformer models \cite{yin_dreaming_2020,ghiasi_plug-inversion_2022}.
 However, it is crucial to note the relationship of memorization, membership inference, inversion and \textit{replication}: A generative model that memorizes data might allow for model inversion or only membership inference, yet the same model might never spontaneously \textit{generate} the training data by accident.

\smallskip
\textbf{Memorization in language.} 
It is well known that generative language models risk replication from their training set
\cite{carlini_extracting_2021,carlini_quantifying_2022} and the amount of replicated data is broadly proportional to the size of the model, amount of duplication of the data point in the training set, and the amount of prompting. Interestingly, such replication behavior occurs even for models that are not overfitting to their training data \cite{tirumala2022memorization,jagielski_measuring_2022}.

\smallskip
\textbf{Diffusion models.} \looseness -1 Diffusion is a process for converting samples from a Gaussian noise distribution into samples from an arbitrary and more complex distribution, such as the distribution of natural images. 

We consider several variants of diffusion models. {\em Stable Diffusion} is a state-of-the-art text-conditional latent diffusion model \cite{rombach2022high}, trained on the LAION database \cite{schuhmann2022laion}. The version we analyze in this work (v1.4) was initially trained on over 2B images and then fine-tuned with 600M images from the LAION Aesthetics v2 $5+$ subset, which is filtered for image quality. We search for matches only in the much smaller 12M LAION Aesthetics v2 $6+$ split to keep storage costs manageable. 

\smallskip
\textbf{Related work.} Replication behavior in GANs has been studied in a number of works.
Meehan et al~\cite{meehan2020non} describe a hypothesis test that discerns whether generated images are on average closer to the training data than a random sample from a hold-out set. Note that this test is at the population level, and is not designed to flag individual instances of replication. 
Feng et al.~\cite{feng2021gans} study the conditions that lead GANs to replicate training data. They look for copies in pixel-space and find that such replications are inversely proportional to dataset complexity and dataset size.
Webster et al~\cite{webster_this_2021} show on face datasets that GANs can occasionally replicate. Interestingly, these models can produce novel images of known identities from the training data without making verbatim copies.
FID scores for ranking GANs favor models that memorize training data \cite{bai_training_2021}, leading toward a search for measures of generalization without memorization \cite{gulrajani_towards_2022}. This includes
``authenticity scores'' that detect replication~\cite{alaa_how_2022}, but only in the form of noisy pixel-by-pixel copies of the training data.   Similarly, authors of large-scale diffusion models have investigated image replication themselves \cite{nichol_dalle_2022}, reducing 
replication through training data de-duplication, and checking for simple nearest-neighbor matches. 

\section{What Counts as Replication?}
There are many different notions of \textit{replication} from creative work, but we will narrow our scope for the purpose of designing a detection system for replicated content. We consider the following (informal) definition:

{\em  We say that a generated image has replicated content if it contains an object (either in the foreground or background) that appears identically in a training image, neglecting minor variations in appearance that could result from data augmentation. }  
 
We focus on object-level similarity because it is likely to be the subject of intellectual property disputes.  We also discount minor differences in appearance that can be explained by data augmentation as these variations would typically not be relevant to a copyright claim. 
An alternative notion is style-wise or semantic similarity.  We do not focus on such definitions here as they are highly subjective, typically are not considered an infringement of intellectual property, and also because many images lack a well-defined style (e.g., natural, unfiltered images from a standard camera).

\section{Detecting Content Replication} 
\label{subsec:retr_setup}
Our goal is to construct a system to detect replication as defined above. To find a powerful system, we consider 10 different prototypes of feature extractors drawn from the SSL and image retrieval literature.  We compare and contrast these methods using 10 different datasets that we curate for measuring the performance of replication detectors. 

\begin{table*}[t]
\centering

\caption{\looseness -1 We present the mAP scores for all 10 models across 10 datasets. The first five datasets are real and the next five are synthetic. In the last column we show the average rank of each model across datasets. We split the models into 3 sub-categories depending on the style of training. The descriptions of the short forms used in the table are:  CD/IR - Copy Detection/ Instance Retrieval, PT - Pre-Trained, SSL - Self-Supervised Learning. Refer \cref{subsec:retr_setup} for more details on models, datasets and the metric. mAP higher the better. Average rank lower the better.}
\label{tab:retrieval_results}
\resizebox{\textwidth}{!}{
\begin{tabular}{@{}llcccccp{0.1\linewidth}p{0.1\linewidth}p{0.1\linewidth}p{0.1\linewidth}p{0.1\linewidth}p{0.1\linewidth}@{}}

\toprule
\textbf{Type} & \textbf{Method}  & rOxford5k $\uparrow$ & rParis6k $\uparrow$ & CUB-200 $\uparrow$ & GPR1200 $\uparrow$ & INSTRE $\uparrow$ & MSCOCO Segmix $\uparrow$ & VOC-Segmix $\uparrow$ & IN-Cutmix $\uparrow$ & IN-Dif-Diagonal $\uparrow$ & IN-Dif-Outpaint $\uparrow$ & Average Rank $\downarrow$\\  
\midrule
\multirow{2}{1ex}{CD/IR}
& Multigrain~\cite{berman2019multigrain}, ResNet-50                         & 22.87     & 44.74    & 3.67   & 37.09   & 56.72  & 27.77         & 25.8       & 67.54     & 79.44         & 24.91     & 5.9    \\
& SSCD~\cite{pizzi2022self}  , ResNet-50                             & 30.16     & 45.75    & 2.00      & 31.42   & 53.54  & 67.04         & 65.82      & 89.78     & 99.91         & 96.11   &\textbf{ 4.2 }      \\

\hline
\multirow{4}{1ex}{PT}
& ViT~\cite{dosovitskiy2020image} S/16, IN1k        & 31.24     & 61.5     & 13.12  & 40.44   & 54.11  & 23.21         & 22.42      & 61.99     & 48.36         & 14.25    & 5.5     \\
& ViT-B/16, IN12k        & 13.14     & 30.43    & 4.24   & 16.63   & 29.15  & 18.15         & 15.61      & 52.74     & 49.69         & 10.18     & 9.2    \\
& ViT-B/16, CLIP~\cite{radford2021learning} on LAION~\cite{schuhmann2022laion}            & 39.92     & 68.92    & 8.6    & 62.13   & 73.19  & 20.37         & 17.91      & 59.5      & 47.54         & 8.76      & 5.4    \\
& Swin Transformer~\cite{liu2021swin}, Base, IN1k      & 40.06     & 72.07    & 15.49  & 54.09   & 68.46  & 24.51         & 24.31      & 74.79     & 40.74         & 14.75    & \textbf{4.1}     \\

\hline
\multirow{4}{1ex}{SSL}
& MoCo~\cite{chen2021empirical}, ViT-B/16                     & 30.25     & 51.6     & 4.94   & 37.98   & 51.88  & 36.41         & 32.9       & 65.98     & 59.12         & 20.61  & 5.1       \\
& MoCo, ViT-B/16 + CutMix~\cite{yun2019cutmix}           & 25.01     & 46.73    & 3.44   & 32.23   & 48.58  & 32.83         & 26.11      & 55.74     & 62.88         & 46.96  & 6.5       \\
& VicRegL~\cite{bardes2022vicregl}, ResNet-50          & 28.4      & 53.79    & 3.02   & 34.95   & 50.98  & 40.58         & 37.76      & 69.74     & 80.02         & 40.93    & 5.0     \\
& DINO~\cite{caron2021emerging}, ViT-B/16, split-product         & 32.14     & 45.43    & 5.76   & 29.41   & 50.06  & 46.42         & 45.29      & 93.53     & 98.92         & 95.86    & \textbf{4.1 }   \\

\bottomrule
\end{tabular}}

\end{table*}


\begin{figure}
    \centering
    \includegraphics[width=\columnwidth]{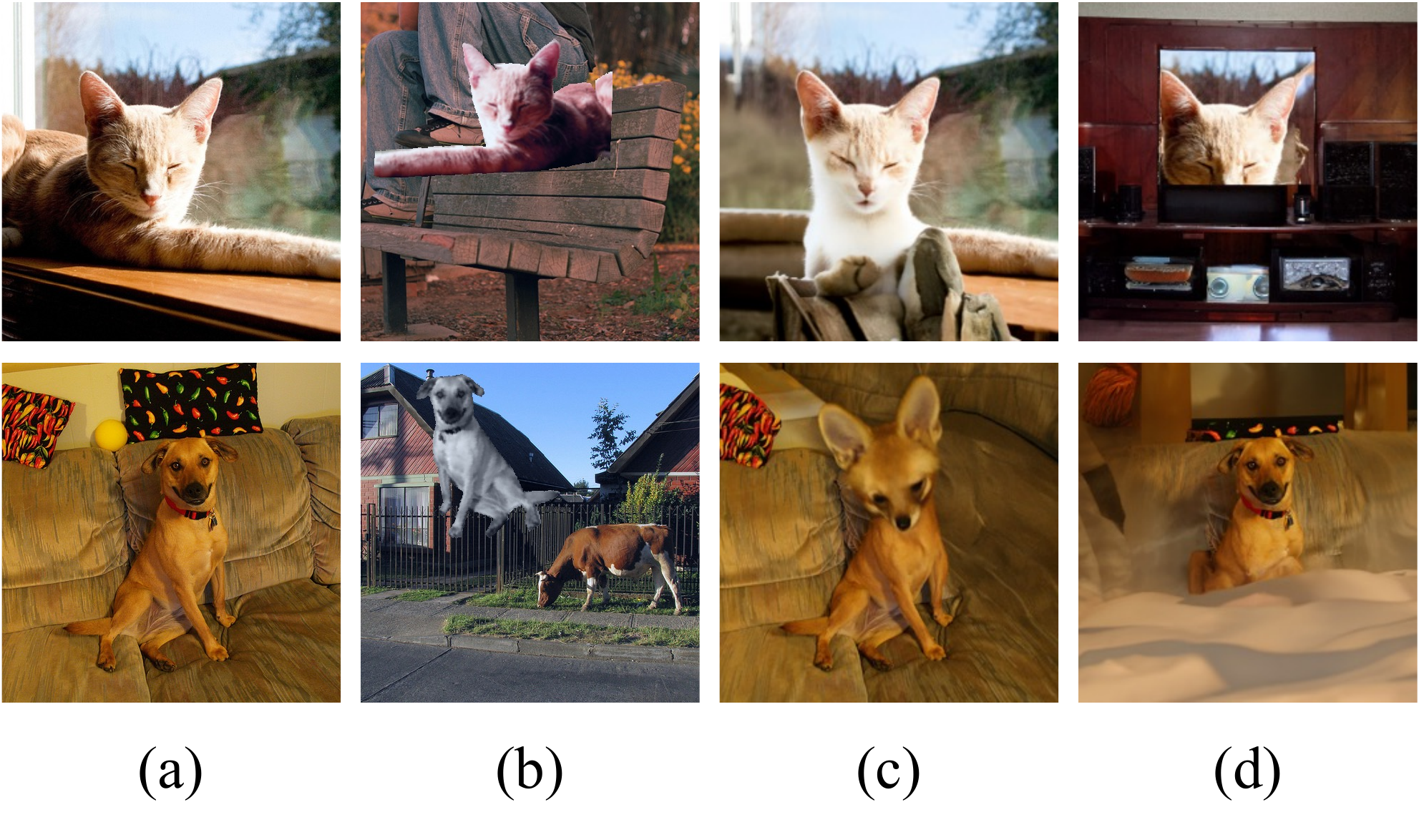} \vspace{-7mm}
    
    \caption{Synthetic datasets. (a) Original images. (b) Segmix generation. (c) Diagonal outpainting. (d) Patch outpainting. Please refer to \cref{subsec:retr_setup} for more details.}
    \label{fig:synthetic_dsets_illustration}
    \vspace{-0.25cm}
\end{figure}

\textbf{Synthetic datasets.} There are currently no existing labeled datasets that capture our notion of replication as defined above. Thus, we create 5 synthetic datasets. Our {\em IN-Cutmix} dataset is built by pasting random square patches from one image into a random location in another. The size of the pasted patch is randomly chosen. We use ImageNet as the source for base images \cite{russakovsky2015imagenet}.  Our {\em IN-Dif-Patch} dataset is created by masking ~80\% of the image except for a random square patch and then outpainting the rest of the pixels using the method proposed by Lugmayr et al~\cite{lugmayr2022repaint}. 

In the above two datasets, the replicated content lies inside a square patch. Vision transformer models naturally rely on square patches, and so we create the {\em  IN-Dif-Diagonal} dataset by masking a random triangular half of an ImageNet image (above or below the diagonal) and using diffusion to outpaint the masked region, resulting in an image that shares half the content of the original.  

Since real world objects may have irregular shapes, we next use the segmentation masks from the MS COCO~\cite{lin2014microsoft} and Pasal VOC~\cite{everingham2015pascal} datasets to generate the synthetic data. For a random query image, we choose either a single object or its background. We then apply a plethora of augmentations (flips, blur, autocontrast, solarize, colorjitter) to the selected region before pasting it into another random image.  If a foreground object is chosen, we also resize and reposition the object at random. We call these challenging datasets {\em MSCOCO Segmix} and {\em VOC Segmix}.  See \cref{fig:synthetic_dsets_illustration}.

\textbf{Real datasets.} Similarity-based image retrieval is closely related to copy detection, although the matching criteria is less stringent for retrieval. We choose 5 image retrieval datasets with high diversity.
 {\em Oxford}~\cite{philbin2007object} and {\em Paris}~\cite{philbin2008lost} are geographic landmark datasets where query and gallery images contain the same building. We use the cleaned-up version with corrected labels ~\cite{radenovic2018revisiting}. {\em INSTRE}~\cite{wang2015instre} contains objects like toys or irregularly-shaped products placed in different locations and conditions. {\em GPR1200} is a general-purpose content retrieval dataset with 1200 classes sampled from other datasets such as Google Landmarks V2~\cite{weyand2020GLDv2}, Stanford Online Products~\cite{oh2016deep}, IMDB-WIKI~\cite{rotheimdb} and others. Caltech-UCSD Birds-200 or {\em CUB-200}~\cite{wah2011caltech} is a dataset with fine-grained classes of birds in different backgrounds, poses and lighting conditions.

\textbf{Models.} Several recent self-supervised (SSL) methods are competitive with supervised retrieval techniques. 
To the best of our knowledge, no rigorous study exists that compares multiple SSL models to retrieval specialist models across multiple datasets. Our study uses the following candidate models and training methodologies. 

MultiGrain~\cite{berman2019multigrain} trains a retrieval model with both classification and retrieval triplet loss. We used the best-performing ImageNet pre-trained ResNet-50 checkpoint from the official repo\footnote{\href{https://github.com/facebookresearch/multigrain}{\texttt{github.com/facebookresearch/multigrain}}}. SSCD~\cite{pizzi2022self} is a self-supervised copy detection method trained in the style of SimCLR~\cite{chen2020simple} using InfoNCE loss~\cite{oord2018representation}, entropy regularization on latent space representations, and many strong augmentations. We used the official ResNet-50 checkpoint trained on ImageNet\footnote{\href{https://github.com/facebookresearch/sscd-copy-detection}{\texttt{github.com/facebookresearch/sscd-copy-detection}}}. Some established methods use pre-trained models as backbones to perform image retrieval tasks~\cite{chen2022deep}. Hence we also evaluate 4 models from timm~\cite{rw2019timm} that are trained in a supervised fashion. ViT-Small/16~\cite{dosovitskiy2020image} pre-trained on ImageNet, ViT-Base/16 pre-trained on ImageNet-21k, ViT-Base/32  image encoder from the CLIP~\cite{radford2021learning} model trained on LAION~\cite{schuhmann2022laion}, and finally a Swin-Transformer~\cite{liu2021swin} with base patch 4 and window 7 trained on ImageNet. 

Lastly, we explored 3 self-supervised models. First is a ViT-Base/16 variant trained with the DINO~\cite{caron2021emerging} framework\footnote{\href{https://github.com/facebookresearch/dino}{\texttt{github.com/facebookresearch/dino}}}.  We also consider ResNet-50 from VICRegL~\cite{bardes2022vicregl}\footnote{\href{https://github.com/facebookresearch/VICRegL}{\texttt{github.com/facebookresearch/VICRegL}}}. Finally we consider ViT-Base/16 from MoCo v3~\cite{chen2021empirical}\footnote{\href{https://github.com/facebookresearch/moco-v3}{\texttt{github.com/facebookresearch/moco-v3}}}, and a variant of MoCo v3 that we fine-tune for 50 epochs with CutMix~\cite{yun2019cutmix} as an additional augmentation with the goal of boosting its copy detection performance.

\textbf{Computing the similarity.} It is common to compare two images via the inner product of feature vectors (either the \texttt{[CLS]} token or average-pooled representations)~\cite{chen2022deep,caron2021emerging, chen2021empirical}. Inner product metrics measure global, rather than local similarity. This is because inner product spaces are metric spaces and thus satisfy the triangle inequality.  To see why this is a problem, consider an example in which generated image $\mathcal{I}_{\text{gen}}$ contains a car and a tree directly stolen from two unrelated images $\mathcal{I}_{\text{car}}$ and $\mathcal{I}_{\text{tree}}$, respectively. Then we would like $d(\mathcal{I}_{\text{gen}}, \mathcal{I}_{\text{car}})$ and $d(\mathcal{I}_{\text{gen}}, \mathcal{I}_{\text{tree}})$ to be very small indicating replication.  But by the triangle inequality, the two unrelated images satisfy $d(\mathcal{I}_{\text{car}}, \mathcal{I}_{\text{tree}}) \leq d(\mathcal{I}_{\text{gen}}, \mathcal{I}_{\text{car}}) + d(\mathcal{I}_{\text{gen}}, \mathcal{I}_{\text{tree}}),$ and are also scored as similar even though they share nothing.

To bypass this potential problem, we implement a \emph{split-product} metric that breaks each feature vector into chunks, computes inner products between corresponding chunks, and returns the maximum across these inner products. In vision transformers, we use the representation corresponding to each token as a chunk since they are more local in nature than the \texttt{[CLS]} token. Under this strategy,  if $d(\mathcal{I}_{\text{gen}}, \mathcal{I}_{\text{car}})$ and $d(\mathcal{I}_{\text{gen}}, \mathcal{I}_{\text{tree}})$ are small, then for each of these two image pairs, at least one such feature vector chunk must yield a high inner product.  However, the locations of these two chunks, each corresponding to one of the image pairs, may differ so that $d(\mathcal{I}_{\text{car}}, \mathcal{I}_{\text{tree}})$ may remain large.  We test both the split-product and standard inner product metric and find that both can return suitable, and often differing, matches.  Qualitatively, the split product metric is a more semantic measure of similarity.  As expected, the inner product metrics enforces a stricter notion of pixel-wise similarity.  See Figure \ref{fig:red}. 

\textbf{Performance metrics.} We measure model performance using mean-Average-Precision or {\em mAP}~\cite{perronnin2009family}. We first compute the average precision for a given query image, and then compute the mean over all queries to find the mAP. Another metric is {\em Recall@k} which measures the proportion of correct matches among the top \textit{k}. {\em MRR} (Mean Reciprocal Rank) records the number of results returned before the first correct match, and computes the mean of the reciprocal of these misses. Higher is better for all metrics.

\subsection{Choosing the Best Replication Detector}
\Cref{tab:retrieval_results} shows mAP scores for all the models across different datasets. We also present average ranks of each model averaged across all datasets (Lower is better). DINO~\cite{caron2021emerging} with split-product performed the best on average across all 10 datasets. For real datasets, the winner is Swin Transformer~\cite{liu2021swin}, and for synthetic datasets SSCD~\cite{pizzi2022self} does best. We also observed similar ranking order when evaluated with other metrics such as MRR, Recall@1, Recall@5, etc (see supplementary material). 

Below, we focus our studies on SSCD, Swin, and DINO (with split-product) as best performing methods in \Cref{tab:retrieval_results}.

\begin{figure}[t]
    \centering
    \includegraphics[width=
    \linewidth]{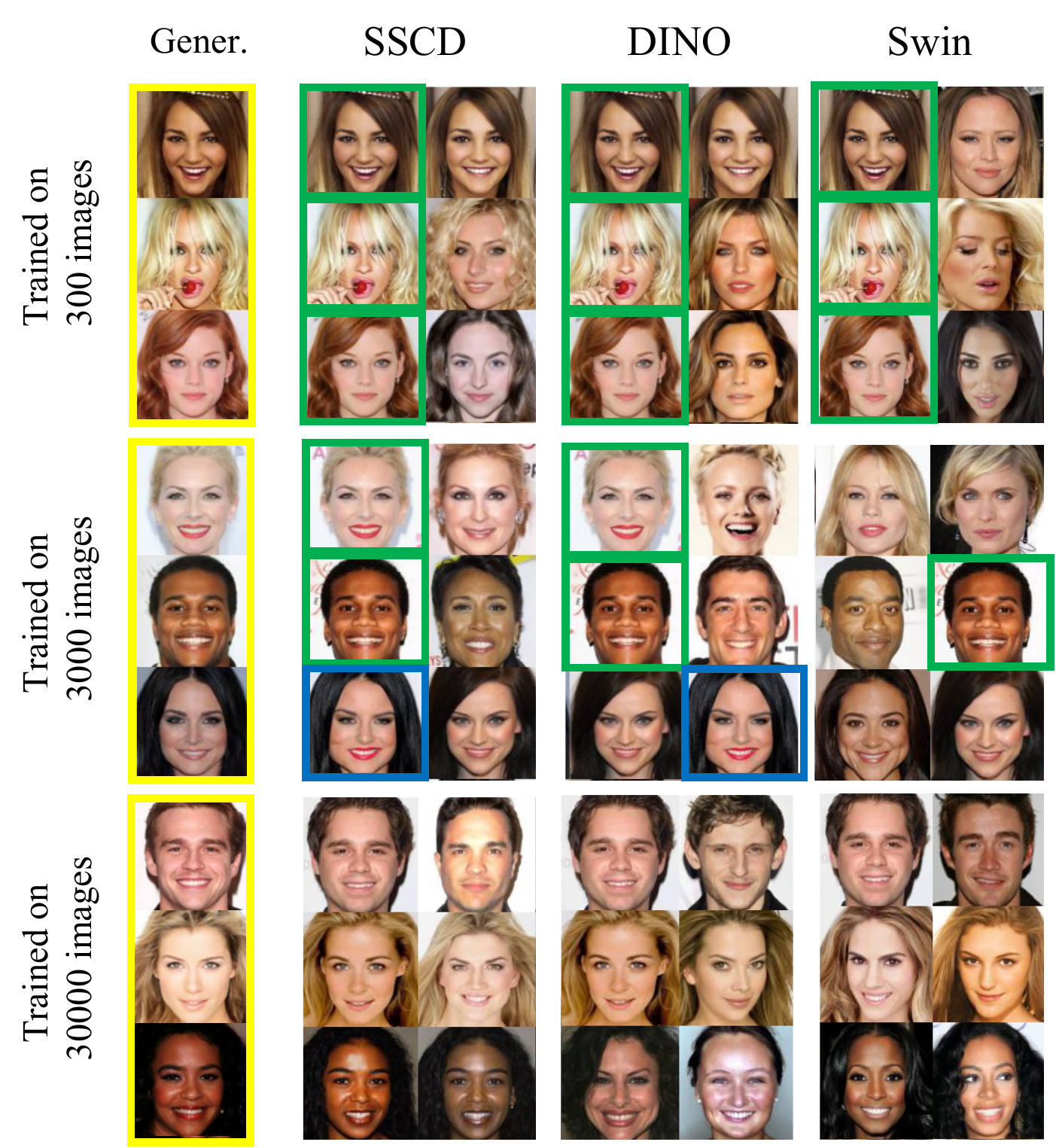}
    \caption{The top two matches (according to different feature extractors) for generations across diffusion models trained on datasets of size 300, 3000 and 30000 (whole dataset). Across the board, we can see full replication in the first 2 models (indicated in green). Very close but not exact copies are indicated in Blue. However in the model trained on the whole dataset, the first matches are \textit{very similar} but not the same. Refer to \cref{sec:ddpm_training_section} for more details about this experiment.}
    \label{fig:celeba_matches}
    \vspace{-.15cm}
\end{figure}

\begin{figure}[ht]
    \centering
    \includegraphics[width=\columnwidth]{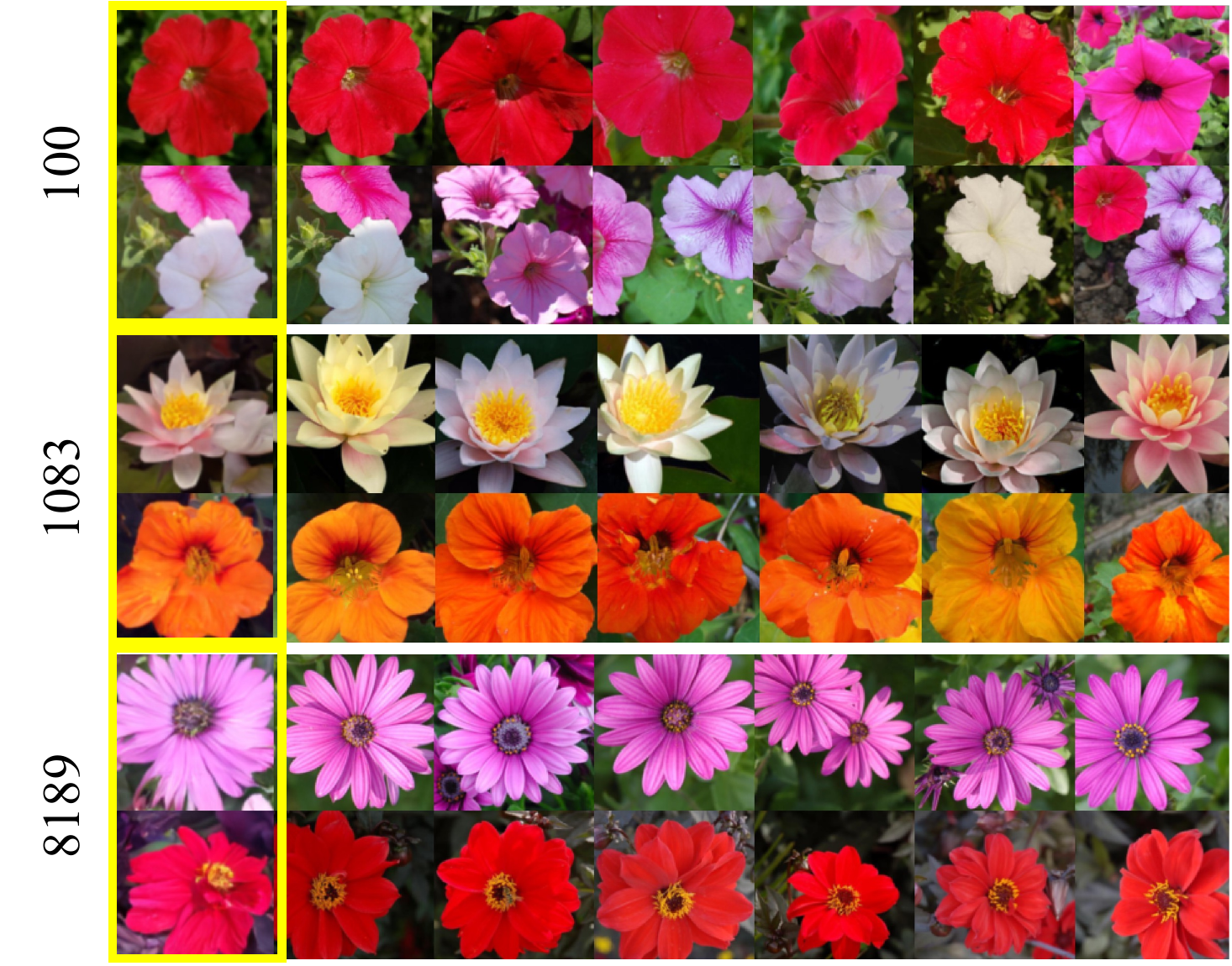}
    \caption{Each row shows the top-6 matches based on the DINO split-product extractor for a given generation. We show two generations from all the three models discussed in \cref{sec:ddpm_training_section}. Full replications decrease as we move from top to bottom.}
    \label{fig:flowers_matches}
    \vspace{-.15cm}
\end{figure}

\section{Do Diffusion Models Copy?}
\label{sec:ddpm_training_section}

In this section, we methodically explore diffusion models trained on different datasets with varying amounts of training data. We observe that the diffusion models trained on smaller datasets tend to generate images that are copied from the training data. The amount of replication reduces as we increase the size of the training set.

\begin{figure*}[ht]
     \centering
     \begin{subfigure}[b]{0.33\linewidth}
         \centering
         \includegraphics[width=\linewidth]{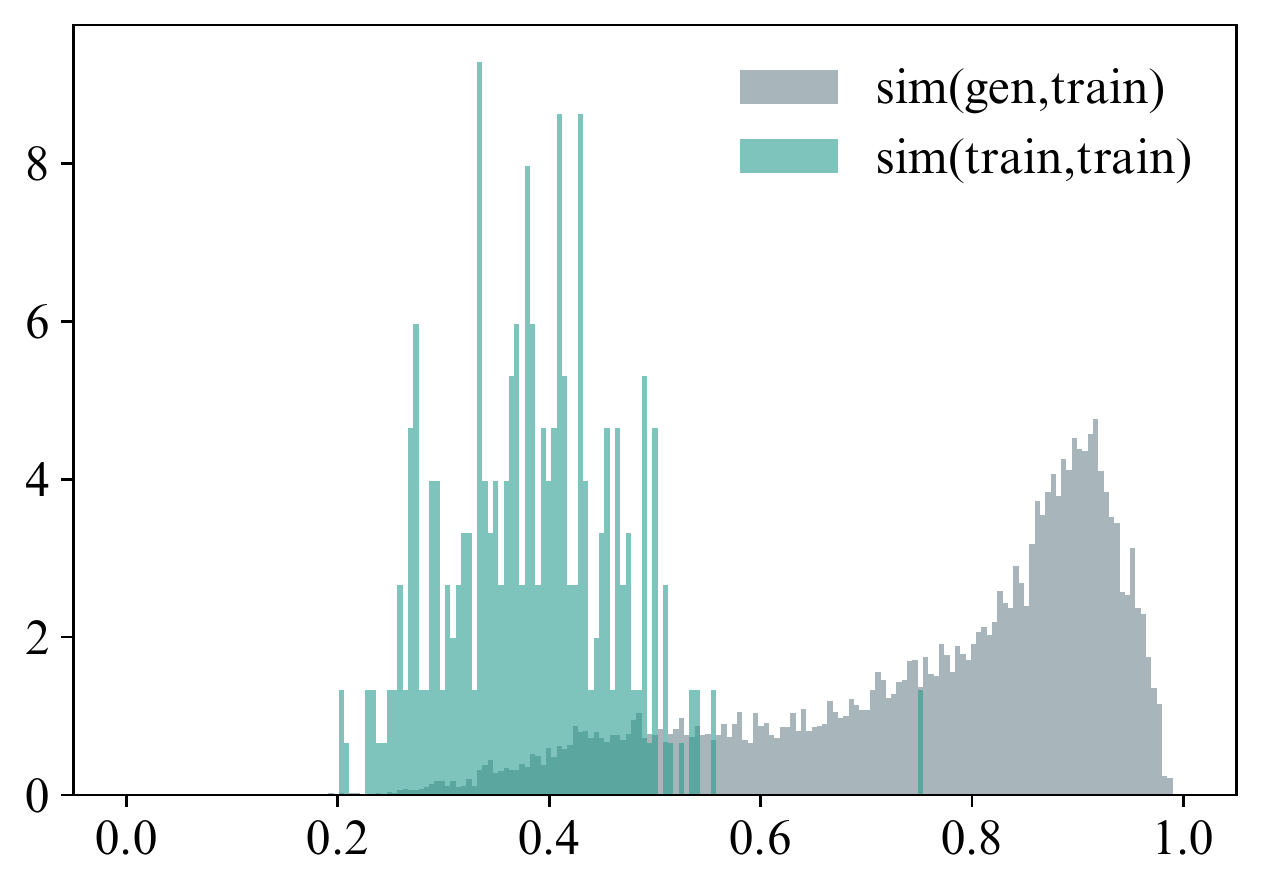}
         \caption{300 points }
         \label{fig:celeba_density_300}
     \end{subfigure}
     \hfill
     \begin{subfigure}[b]{0.33\linewidth}
         \centering
         \includegraphics[width=\linewidth]{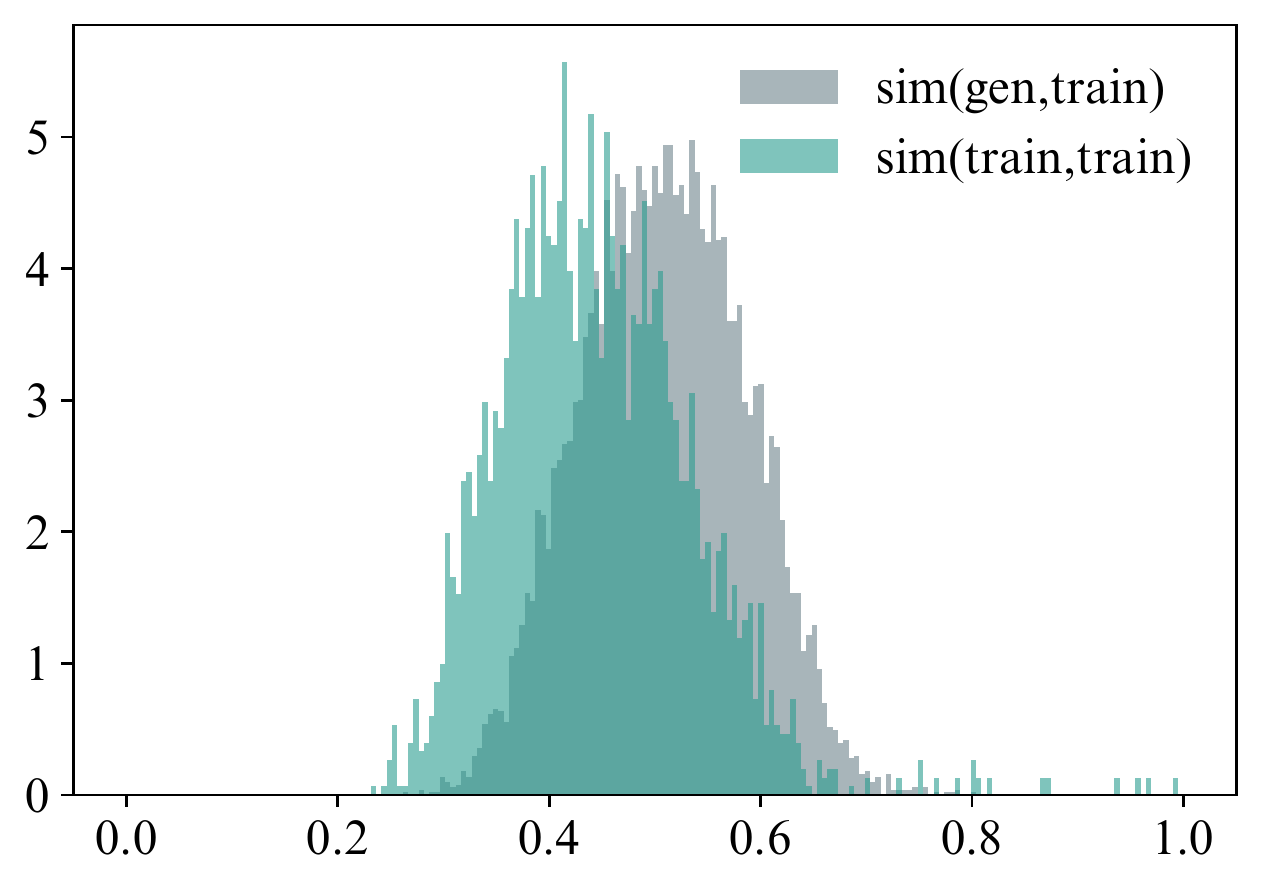}
         \caption{3000 points}
         \label{fig:celeba_density_3000}
     \end{subfigure}
     \hfill
     \begin{subfigure}[b]{0.33\linewidth}
         \centering
         \includegraphics[width=\linewidth]{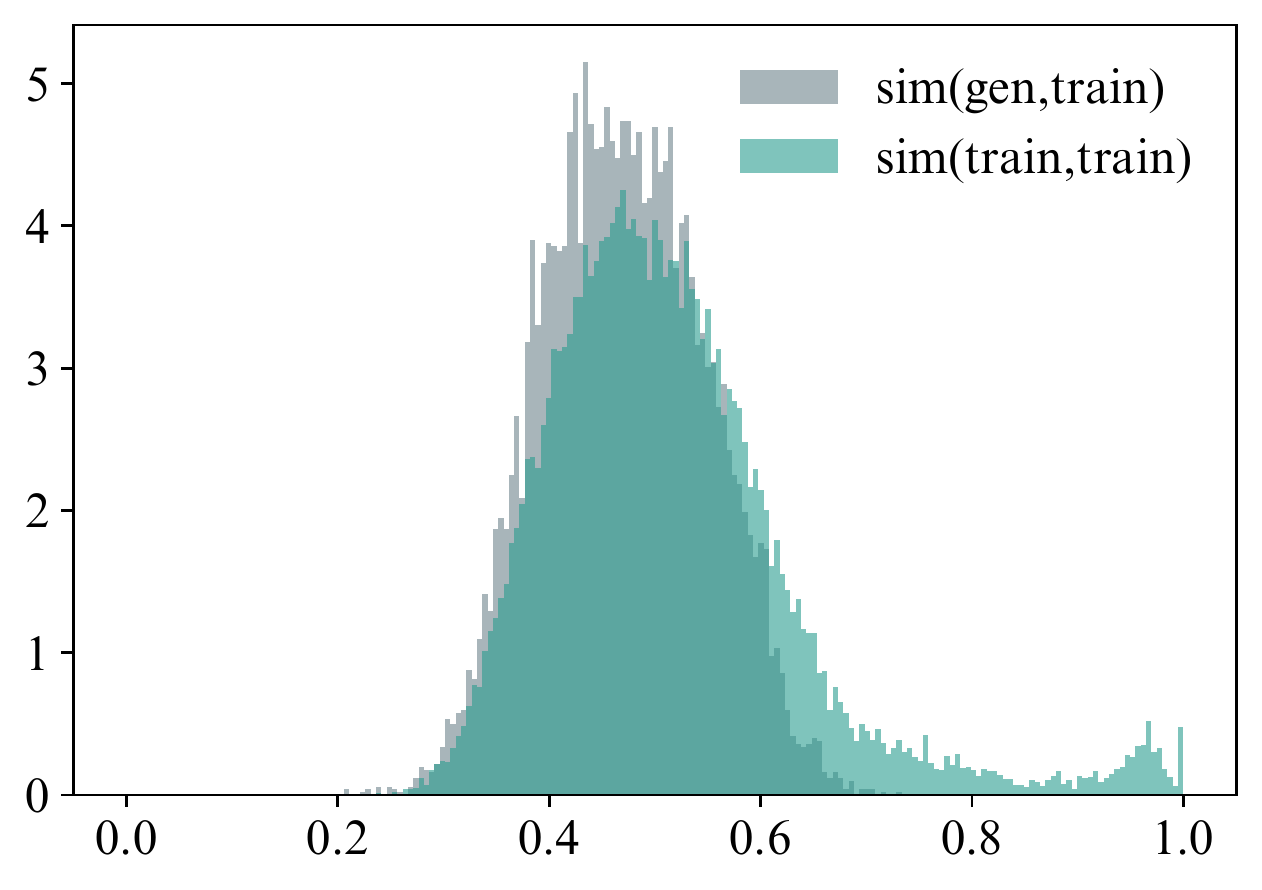}
         \caption{All points}
         \label{fig:celeba_density_all}
     \end{subfigure}
     \captionsetup{font=small}
        \caption{The histograms of top-1 similarity scores between generations and the training data and the top-1 self-similarity scores of training data across 3 diffusion models trained with different amounts of training data. Varying amount of training data is used in each plot but $10000$ generations are used across all. }
        \label{fig:celeba_density_plots}
        \vspace{-.25cm}
\end{figure*}

\textbf{Experimental setup.} We train Denoising Diffusion Probabilistic Models (DDPM)~\cite{ho2020denoising} with a discrete denoising scheduler on various datasets using the HuggingFace implementation\footnote{\href{https://huggingface.co/docs/diffusers/api/pipelines/ddpm}{\texttt{huggingface.co/docs/diffusers/api/pipelines/ddpm}}}. For Celeb-A~\cite{liu2015faceattributes}, we train two models on $300$ and $3000$ training images. We also use the full dataset pre-trained checkpoint from the official repository\footnote{\href{https://huggingface.co/google/ddpm-celebahq-256}{\texttt{huggingface.co/google/ddpm-celebahq-256}}}. For Oxford Flowers~\cite{Nilsback08}, we train models on $100$, $1083$ (top $5$ classes), and $8189$ (complete dataset) images. We train all models with random horizontal flip and random crop augmentations. In all cases, we train the models until generations appear to be high quality, for at least $300$k steps and until the FID~\cite{szegedy2015rethinking} scores are lower than $50$. We do quantitative analysis on $10000$ generations for Celeb-A and $5000$ generations for Oxford Flowers. 

\textbf{Finding matches.} For each generated image, we search the training set using dot products between its features and training samples (except for DINO which uses split-product).  All the generations used in figures are from the $20$ generated images with highest top-1 similarity scores for standard diffusion models, and are among images with similarity $>0.5$ for {\em Stable Diffusion}.

\textbf{Qualitative observations.} \Cref{fig:celeba_matches} and \Cref{fig:flowers_matches} show generated images and their corresponding top matches from the training dataset. We consider diffusion models (DDPM) trained with varying amounts of training data. 
In the case of Celeb-A, diffusion models trained on $300$ and $3000$ images blatantly copy from their training images. However, when the model is trained on the whole dataset, generations may appear that are similar to training samples, but not identical. We observe similar trends in diffusion models trained on the Oxford Flowers dataset as well. Curiously, the model trained on just $1083$ images is already able to create unique (but minor) variations of the training data. Note that the samples in \cref{fig:celeba_matches} and \cref{fig:flowers_matches} are not cherry-picked, but they are chosen from the small subset of $20$ images with highest top-1 similarity to the training data.

\textbf{Quantitative observations.} 
To further complement our visual inspection, we can also examine the distribution of similarity scores between generated images and training samples.  \Cref{fig:celeba_density_plots} contains histograms of similarity scores between generations and their best match from the training data. As a baseline, we also draw random training images and compute the similarity with their closest match from the remaining training images.  If most scores between generated and training images lie to the right of this baseline, then the model is generating images that are closer to their training samples than the training samples are to each other.

Most samples generated by the $300$-sample model are extremely similar to the training data, having very high similarity scores. However, the histogram's mass shifts drastically to left when we train the model instead on $3000$ points. We do see blatant copies from this model too, but this phenomenon occurs infrequently. The histograms of similarity scores computed using the full dataset model are highly overlapping. This strong alignment indicates that the model is not, on average, copying its training images any more than its training images are copies of each other.  The histogram of generated images (gray) no longer has a long right tail, indicating that the model is unlikely to generate exact copies of its training samples.  
Note that a small proportion of the dataset self-similarity scores in \Cref{fig:celeba_density_plots}~(c) are greater than $0.9$, indicating that there are repetitions or near repetitions in the training data.

\section{Case Study: ImageNet LDM}
\label{sec:imagenet_ldm}

\begin{figure*}[t]
    \centering
    \includegraphics[width=\linewidth]{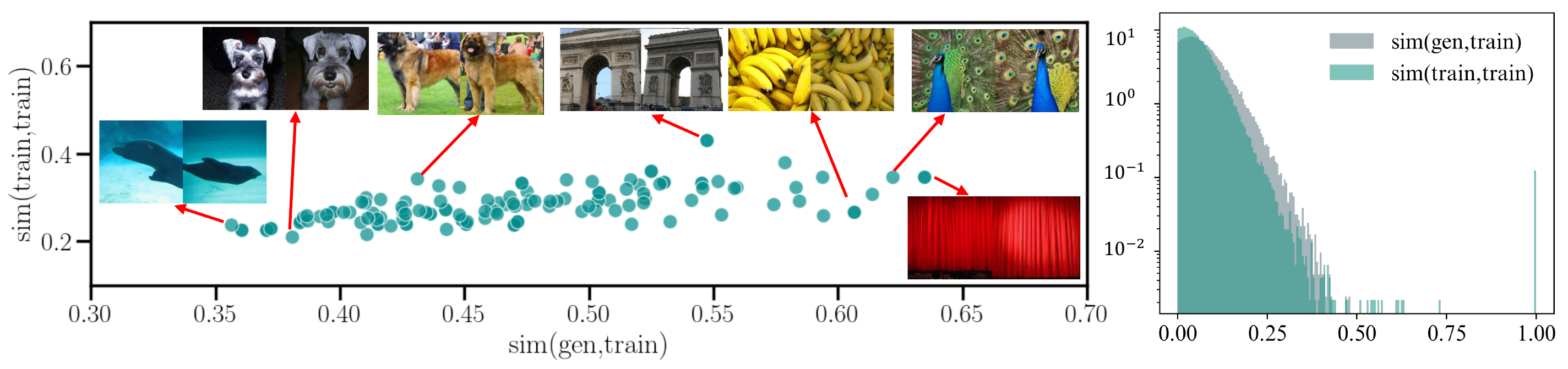} \vspace{-5mm}
    
    \caption{(Left) For each individual class, we plot top-1 similarity between a generation and training data vs the mean of the top-1 similarity scores of the training data to itself. We also show a few images for some interesting classes. Left image is generation and right image is closest match. (Right) We show histograms of top-1 similarity scores for all the generations in $100$ classes and the self-similarity scores of training data (within the class). }
    \label{fig:imagenet_scores_wimages}
\end{figure*}
\begin{figure*}[ht]
    \centering
    \includegraphics[width=\linewidth,trim = 6mm 1mm 0mm 2mm, clip]{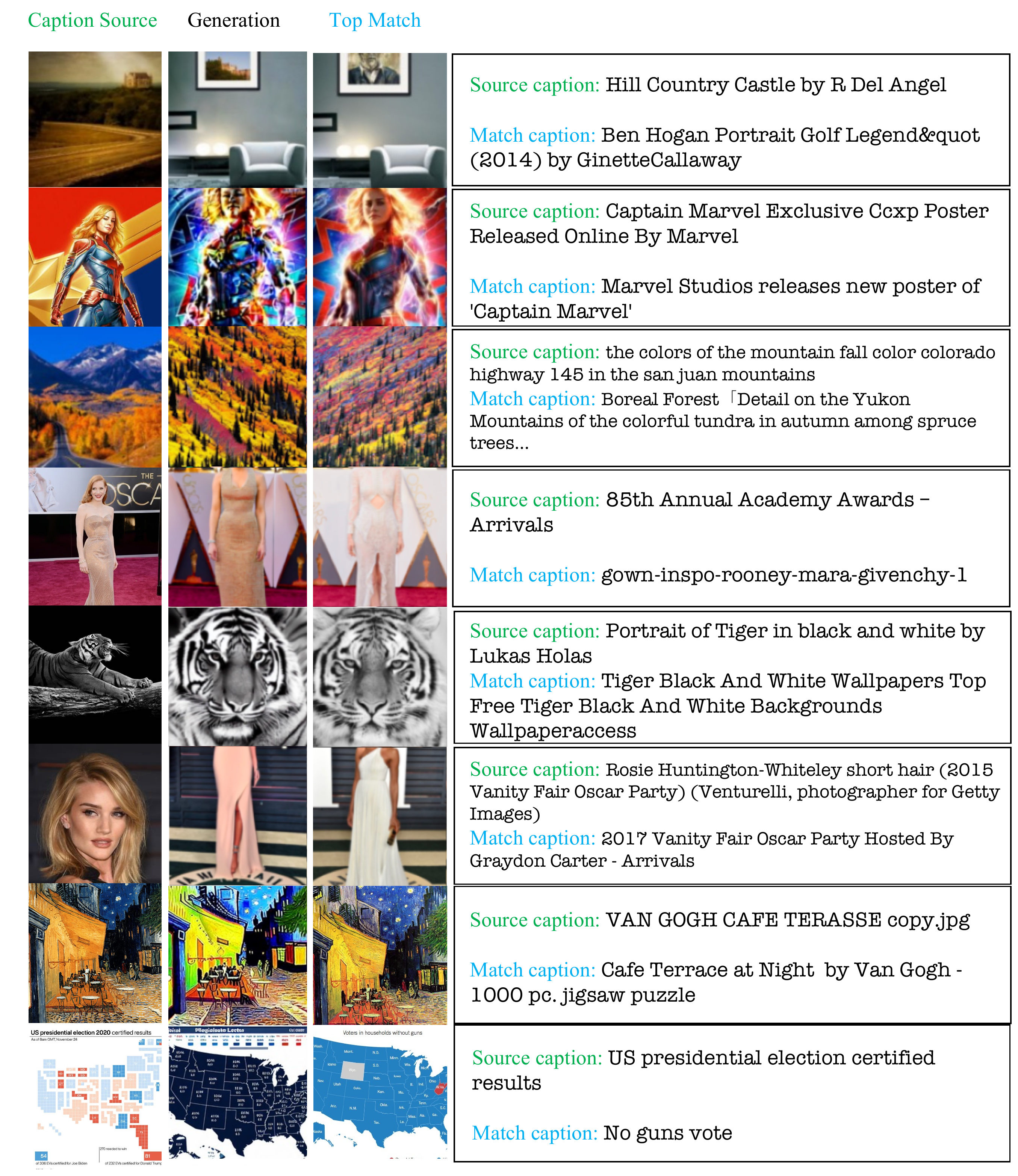}
    \caption{Selected {\em Stable Diffusion} generations using captions sampled from LAION images, with similarity score $\ge 0.5$.}
    \label{fig:sd_train_prompts_copies}
    \vspace{2mm}
\end{figure*}

\textbf{Experimental setup.} In the previous section, we observed copying behavior when diffusion models are trained on small datasets, and the rate of copying decreases as models are trained on more data. In this section, we extend our study to an off-the-shelf class conditional Latent Diffusion Model~\cite{rombach2022high} trained on ImageNet. We search for copying both at the class level and the population level. We use the pre-trained checkpoint from the official repo.\footnote{\href{https://github.com/CompVis/stable-diffusion}{\texttt{github.com/CompVis/stable-diffusion}}} We randomly choose $100$ classes for this study and then generate $1000$ samples per class (comparable to the size of training data per class in ImageNet).

\textbf{Observations.} Qualitatively, we observe no significant copying in any of the generations by this model. In \Cref{fig:imagenet_scores_wimages} (a), we present a scatter plot with x-axis showing the maximum similarity scores observed between generations of a given class and ImageNet training samples. On the y-axis, we show the average similarity scores per class observed between training samples in that class.  For a few interesting points, we also show the corresponding generation and the top match in the training data.  We see the similarity scores never cross $0.65$, and when we manually sift through the high similarity score examples in each of the $100$ classes, they are \textbf{very similar} but never exact copies, and may be explained by low intra-class diversity

We also check if there is a relationship between the intra-class diversity and similarity scores, and indeed classes with higher self-similarity scores on average have higher maximum similarity score amongst matches with generated samples. Specifically, the points in the scatter plot have a correlation of $0.6$ and the line of best fit has slope $0.39$.  The classes with the highest similarity between generated images and training data are \texttt{theater curtain}, \texttt{peacock}, and \texttt{bananas}. Meanwhile \texttt{sea lion}, \texttt{shopping cart}, \texttt{bee}, and \texttt{swing} are at the lower end of the spectrum.
In \Cref{fig:imagenet_scores_wimages} (b), we consolidate our results across the classes into a histogram of similarity scores between generations and matches from the training set and the similarity scores of training images with matches from the remaining training samples.  The average similarity scores are relatively low for this dataset as well as for this diffusion model showing that the chance of replication is very low.

\section{Case Study: {\em Stable Diffusion}}
\label{sec:stable}
In this section, we evaluate {\em Stable Diffusion} v.$1.4$~\cite{rombach2022high}, which was trained on the publicly available LAION~\cite{schuhmann2022laion} dataset. Since it is outside the computational reach of our meager academic cluster to store and search $2$ billion+ images, we narrow our search scope to the smaller LAION Aesthetics v2 $6+$ dataset which has $12$M images and is a \textbf{subset} of images that were used for the final rounds of training. We load the model and the checkpoints via HuggingFace \footnote{\href{https://huggingface.co/CompVis/stable-diffusion-v1-4}{\texttt{huggingface.co/CompVis/stable-diffusion-v1-4}}}. 

In the first experiment, we randomly sample $9000$ images, which we call \emph{source images}, from LAION Aesthetics 12M and retrieve the corresponding captions. These source images provide us with a large pool of random captions.  Then, we generate synthetic images by passing those captions into {\em Stable Diffusion}.  We study the top-1 matches, which we call \emph{match images}, for each generated sample. See the supplementary material for all the prompts used to generate the images for figures as well as the analysis in this section.

We attempt to answer the following questions in this analysis. 1) Is there copying in the generations? 2) If yes, what kind of copying?  3) Does a caption sampled from the training set produce an image that matches its original source? 4) Is content replication behavior associated with training images that have many replications in the dataset?

\begin{figure}
    \centering
    \hspace{0mm} Generated \hspace{7mm} DINO 1 \hspace{6mm} DINO 10 \hspace{4mm} SSCD  1-10\\
    \includegraphics[width=\linewidth]{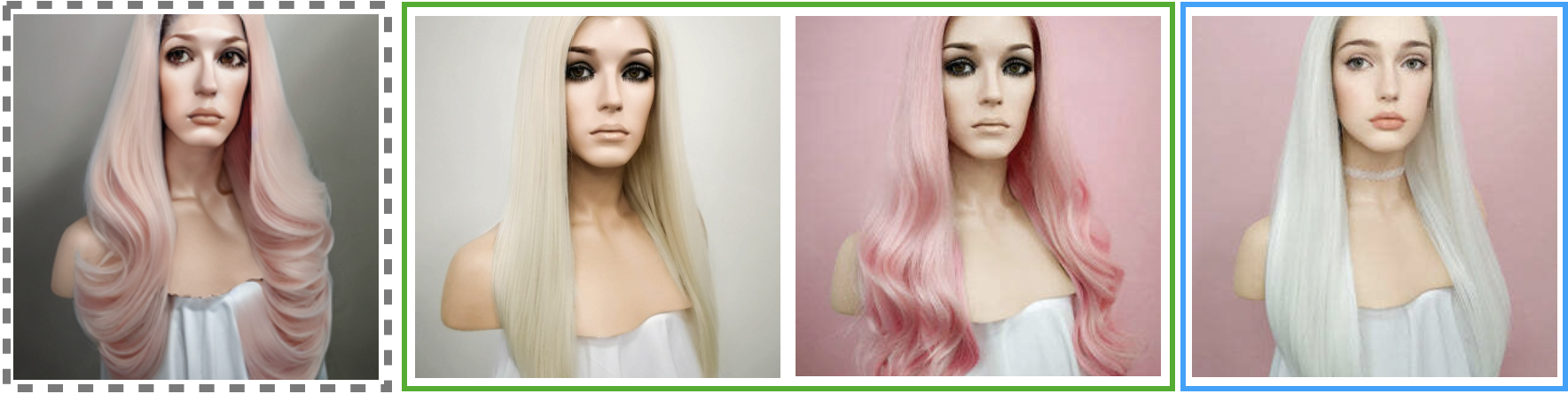}  \vspace{-.5cm}
    \caption{The top-10 SSCD matches for this generation are identical copies of the same training image, and edges (see the shoulder and height of the dress) overlap strictly with the generation.  In contrast, DINO focuses on semantic, rather than pixel-level, similarity and finds a more diverse range of matches.}
    \label{fig:red}
    \vspace{-.2cm}
\end{figure}

In previous experiments, we observed that DINO with split-product is slightly better than SSCD at finding copies. Here, we find that SSCD and DINO often flag the same generated images, but DINO is capable of finding more diverse matches with lower pixel similarity (Figure \ref{fig:red}). 
We constructed visualizations in this section by choosing from images with an SSCD similarity $>0.5$.  

\begin{figure*}
    \centering
    \includegraphics[width=\linewidth, trim = 4mm 4mm 3mm 4mm, clip]{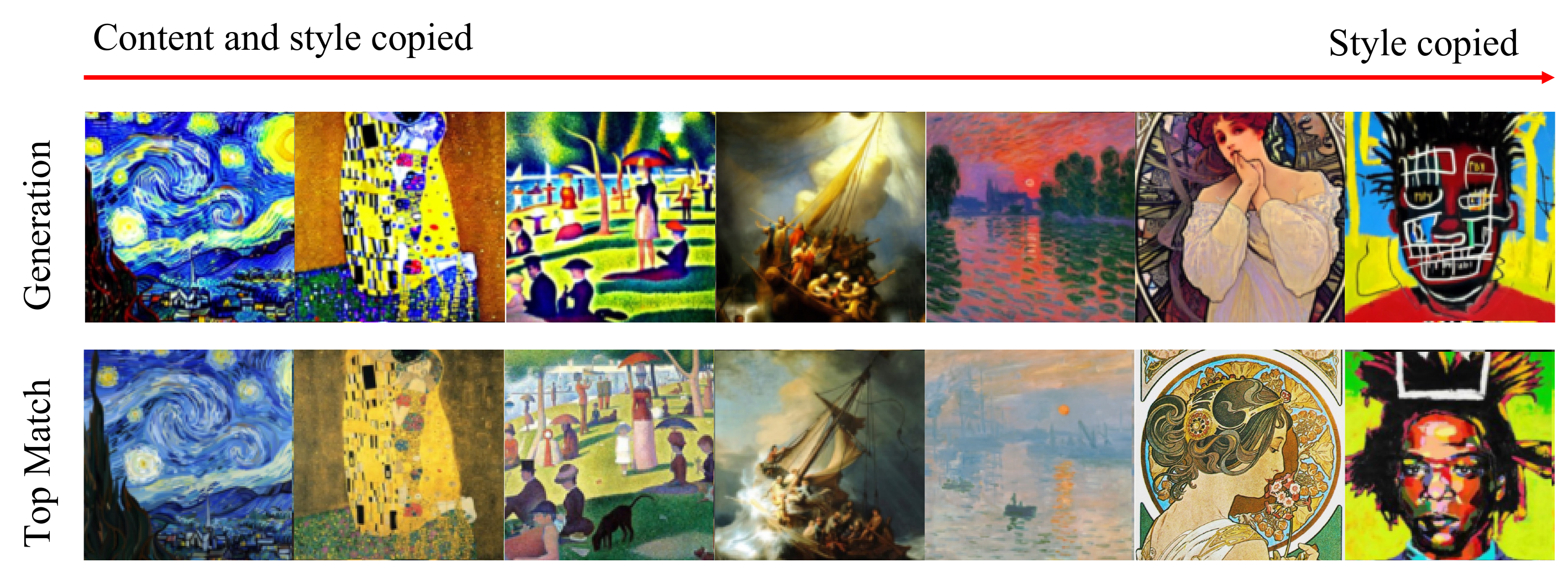}  \vspace{-.6cm}
    
    \caption{ {\em Stable Diffusion} replicates pixel-level details, structures, and styles of well known paintings.}
    \label{fig:sd_paintings_copies}
    \vspace{-.2cm}
\end{figure*}

\textbf{Observations.} In \Cref{fig:sd_train_prompts_copies}, we visualize instances of copying found in samples generated by {\em Stable Diffusion}. We choose them from a small set of points ($\approx 170$ images) whose top-1 similarity scores are $ > 0.5$ (top $1.88$ percentile). Above this $0.5$ threshold, we observe a significant amount of copying. The first row (where only the painting changed) shows verbatim usage of an object and background. The \nth{4} and \nth{6} rows show local copying where only the background is recycled from the training set. We see similar trends in other images with high similarity scores.

While all synthetic images were generated using captions sourced from LAION, none of the generations match their respective source image. 

In fact, sometimes the caption of the source image is not representative of the source image content, and the generation is quite different from the source. This behavior can be seen in the first row of \Cref{fig:sd_train_prompts_copies}. In the fifth row, the match image is more representative of the source caption than the source itself.

\begin{figure}
    \centering
    \includegraphics[width=\linewidth]{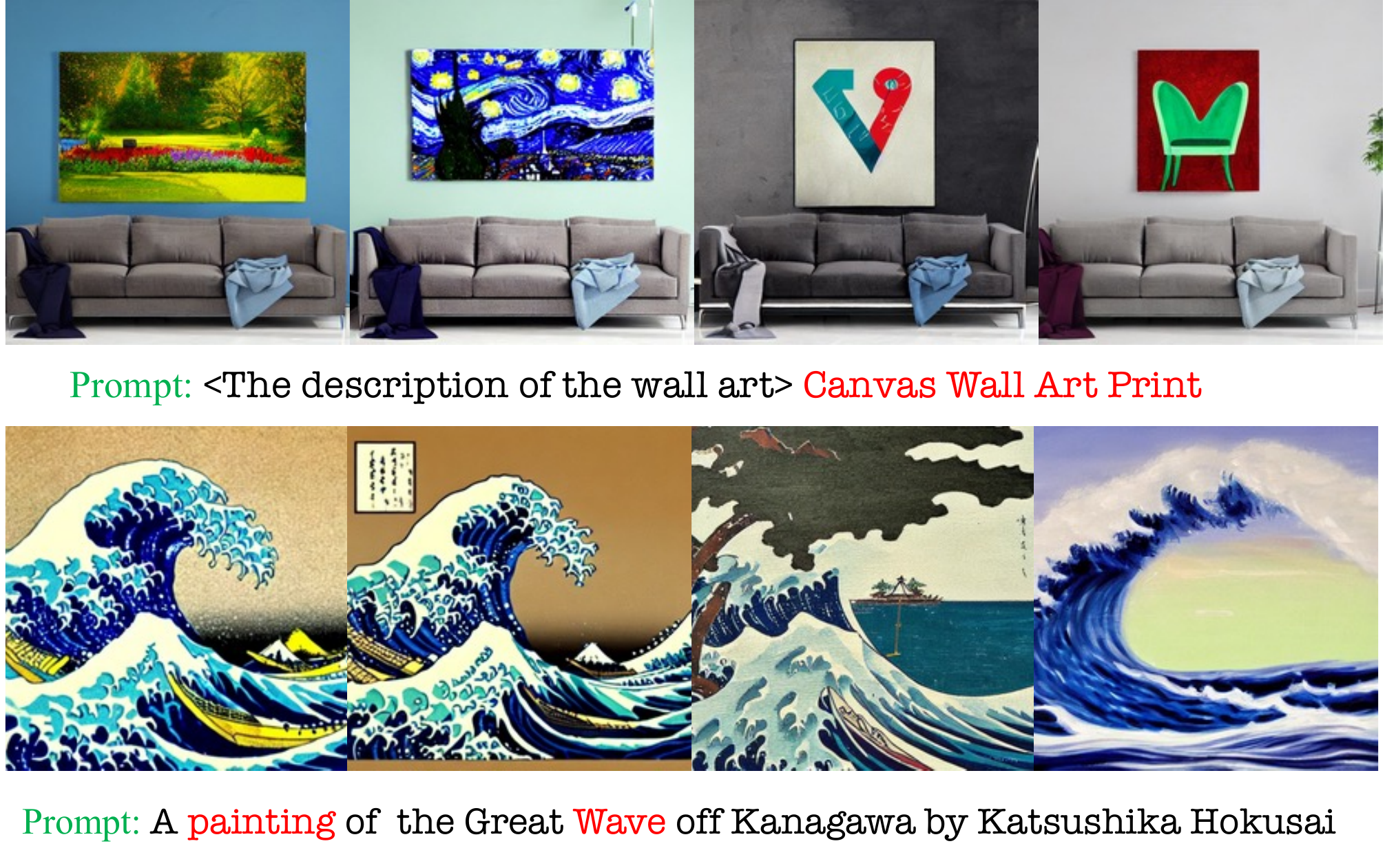}\vspace{-2mm}
    
    \caption{Including the phrase highlighted in red into a random prompt for {\em Stable Diffusion} leads to exact replications of the sofa (top row) and wave shape (bottom row).}
    \label{fig:sd_keyphrase_replic}
    \vspace{-2mm}
\end{figure}

\looseness -1 In those $170$ images, we find instances where replication behavior is highly dependent on key phrases in the caption. We show two examples in \Cref{fig:sd_keyphrase_replic} and highlight the key phrase in red. For the first row, the presence of the text \texttt{Canvas Wall Art Print} frequently ($\approx~20\%$~of the time) results in generations containing a particular sofa from LAION (also see Fig \ref{fig:teaser}). Similarly, the second row shows various generations by tweaking the prompt \texttt{A painting of the Great Wave off Kanagawa by Katsushika Hokusai}. We gradually remove words until only \texttt{painting} and \texttt{wave} remain. All of the generations have a wave structure that resembles the original painting. 

We also notice instances of generations where style is copied rather than content. This can be explicitly seen when the name of an artist is used in the generation prompt. We generate many paintings with the prompt style ``\texttt{<Name of the painting>} by \texttt{<Name of the artist>}''. We tried around $20$ classical and contemporary artists, and we observe that the generations frequently reproduce known paintings with varying degrees of accuracy. In \Cref{fig:sd_paintings_copies}, as we go from left to right, we see that content copying is reduced, however, style copying is still prevalent. We refer the reader to the appendix for the exact prompts used to generate \cref{fig:sd_keyphrase_replic} and \cref{fig:sd_paintings_copies}.
\begin{figure*}[ht]
     \centering
    \begin{subfigure}[b]{0.33\linewidth}
         \centering
         \includegraphics[width=\linewidth]{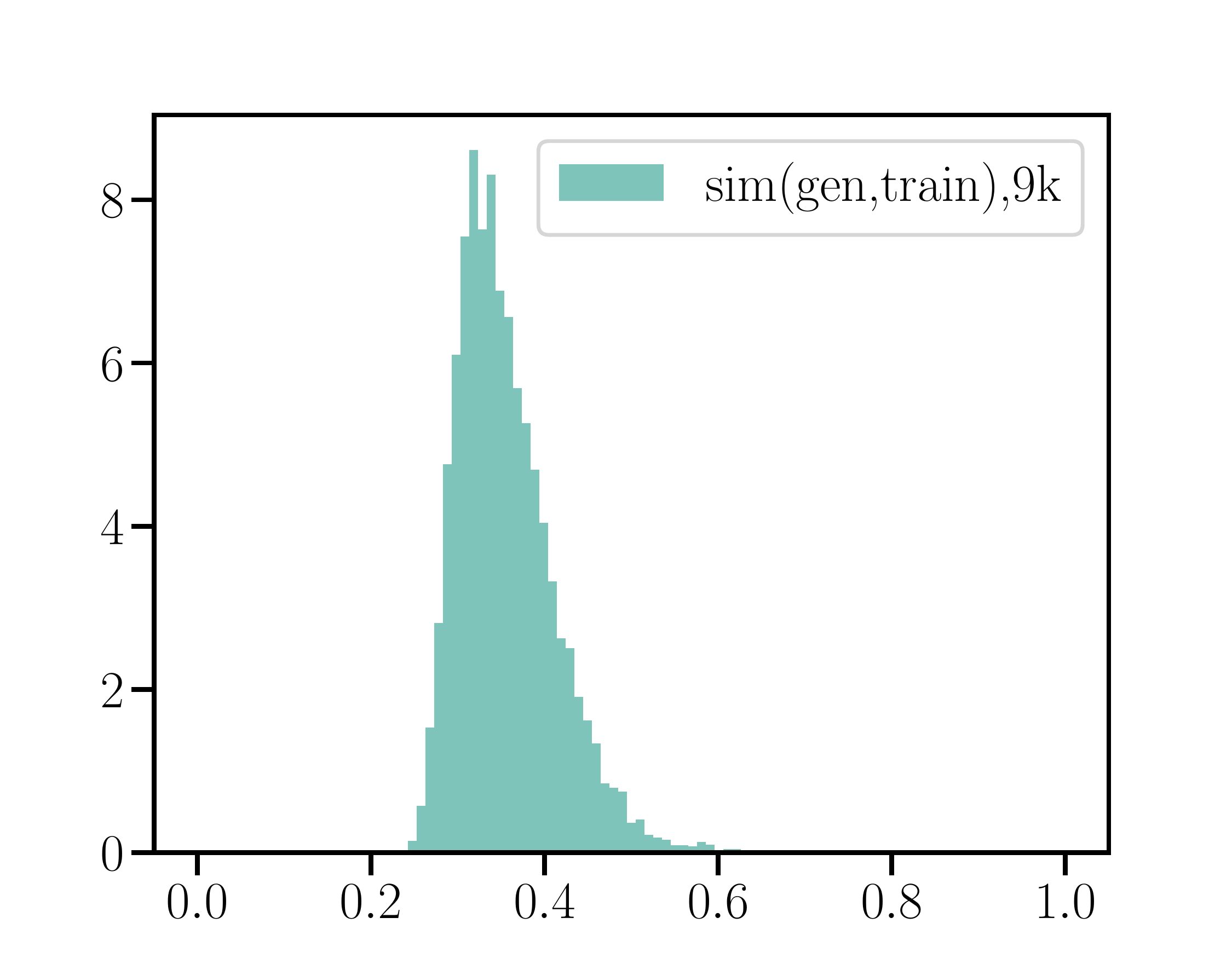}
     \end{subfigure}
     \hfill
     \begin{subfigure}[b]{0.33\linewidth}
         \centering
         \includegraphics[width=\linewidth]{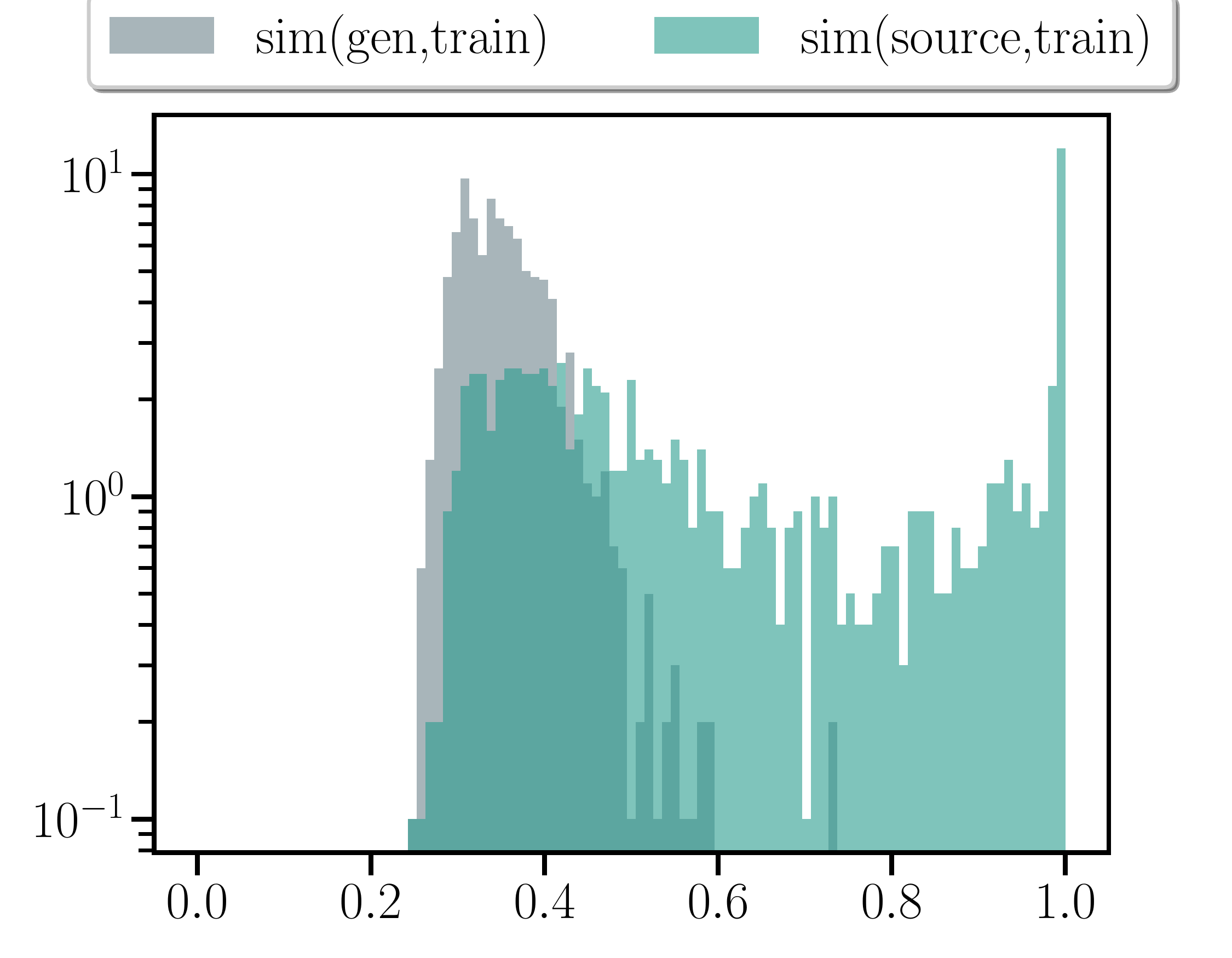}
     \end{subfigure}
     \hfill
     \begin{subfigure}[b]{0.33\linewidth}
         \centering
         \includegraphics[width=\linewidth]{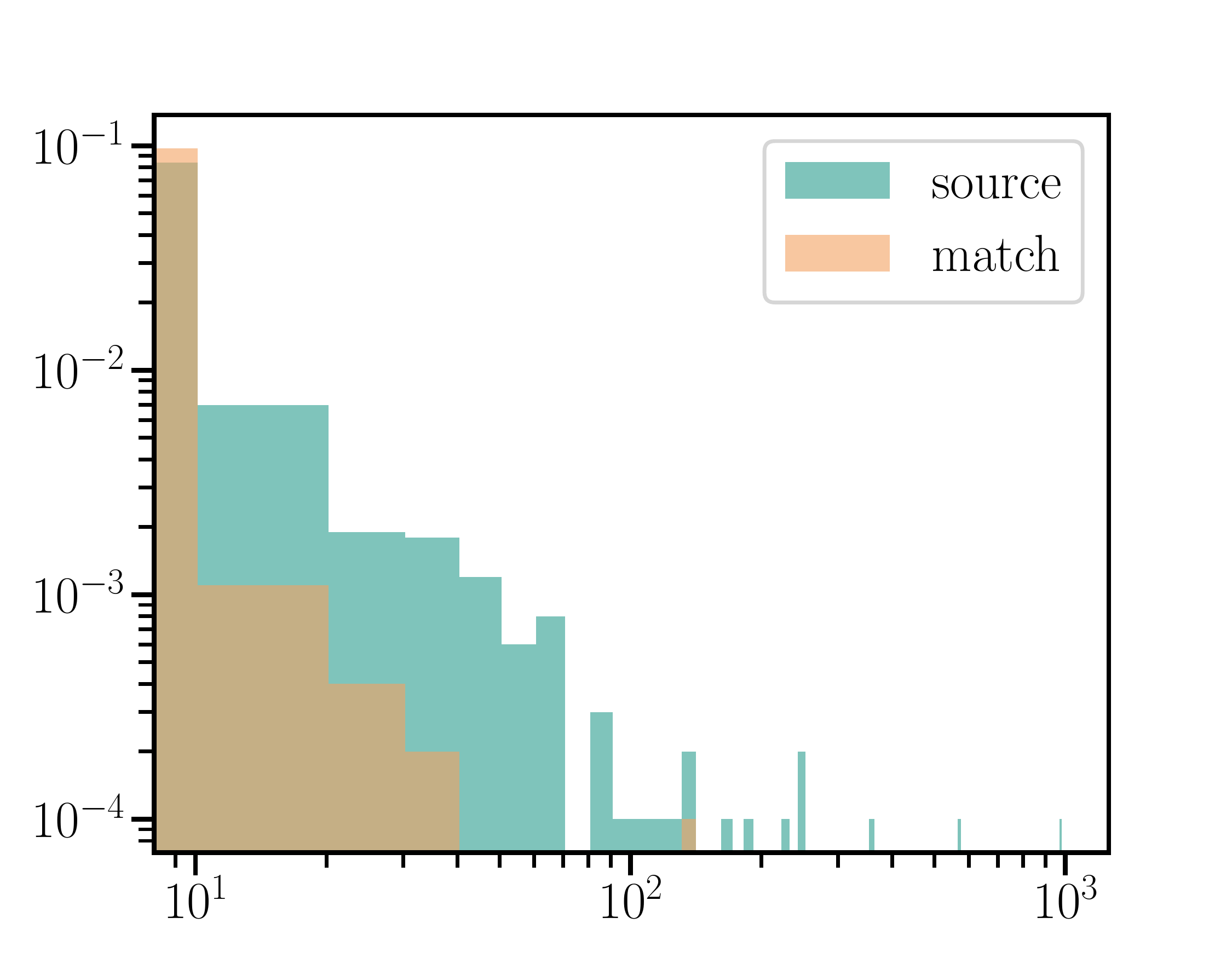}
     \end{subfigure}

        \caption{{\em Stable Diffusion} density plots. (Left) Similarity scores of all 9000 generations w.r.t. train data. (Center) Similarity scores of 1000 random generations with train data and the corresponding 1000 caption source images with train data. (Right) Histogram of number of training set duplications for $1000$ random source images and $1000$ match images. Random data (source) is relatively repeated more than matched images. }
        \label{fig:sd_sim_plots}
\end{figure*}
\begin{figure}
    \centering
    \includegraphics[width=\linewidth]{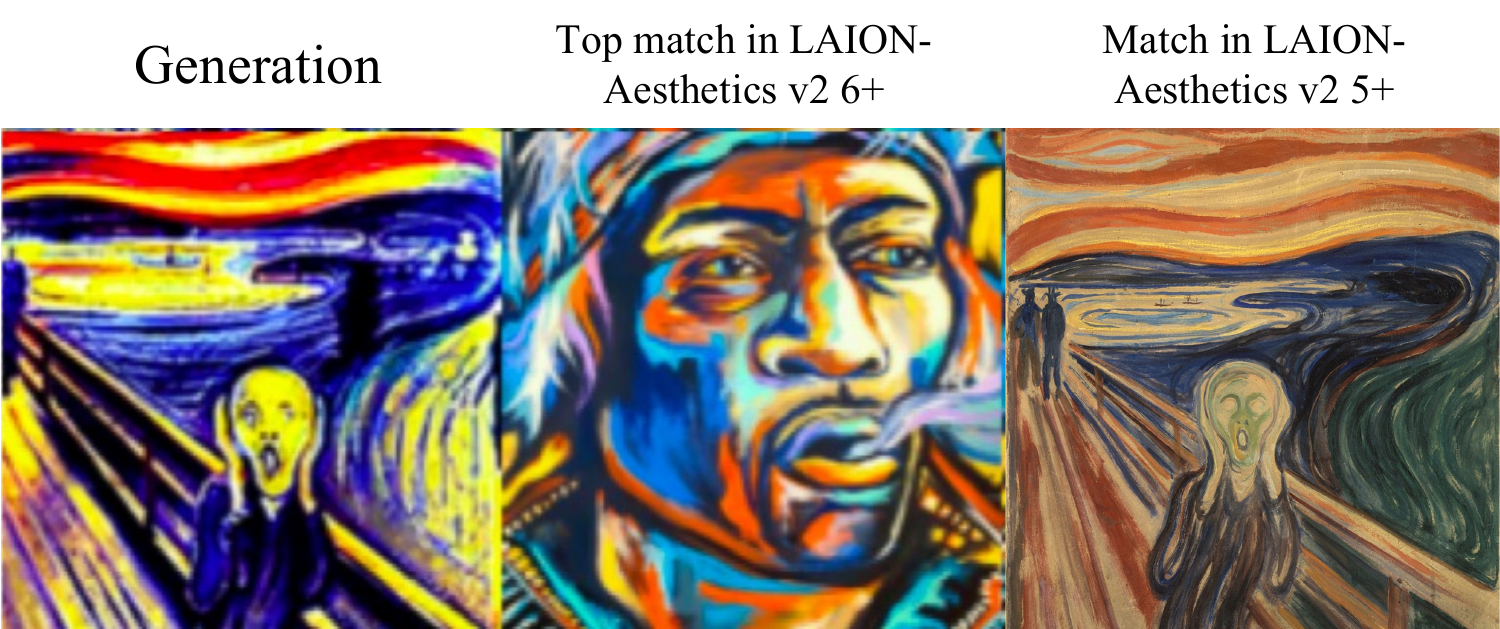}
    \caption{{\em Stable Diffusion} generates the painting ``The Scream." This image is within the 600M image LAION-Aesthetics-$5+$ split that was used for training, but is not within the 12M image LAION-Aesthetics-$6+$ split that we searched in this study.    
    }
    \label{fig:scream_failure}
   \vspace{-.25cm}
\end{figure}

\textbf{Role of caption sampling.}
Several of our studies sample random captions from LAION itself, and this may lead to a replica of the caption source image.
\Cref{fig:sd_sim_plots} (left) shows histograms of top-1 similarity scores for $1000$ random generations, and the top-1 similarity scores of the $1000$ corresponding images from which we sourced the captions. We find the mean generation similarity scores to be much lower than those of the source images, indicating that this method of caption sampling does not typically produce a replica of the caption source.

Still, captions from LAION will sometimes contain ``key phrases'' that are closely associated with dataset images and therefore conjure a memorized image.  For this reason, LAION-based caption sampling may lead to higher rates of data replication than other sampling methods.  It is difficult to correct this bias in a scientific way, as there is no baseline sampler for ``typical'' captions. Furthermore, captions used by experienced diffusion users will often exploit powerful key phrases (e.g. ``art station,'' ``35mm,'' or the name of an artist) that widely appear in LAION.


\textbf{Role of duplicate training data.}
Many LAION training images appear in the dataset multiple times.  It is natural to suspect that duplicated images have a higher probability of being reproduced by {\em Stable Diffusion}.
\Cref{fig:sd_sim_plots} (right) shows a histogram of how many times a training image is duplicated in LAION-Aesthetics, where ``duplicate'' is defined as having SSCD score $>0.95$. We plot a histogram for two populations. First, the 1000 uniformly sampled source images.  Second, we generate 1000 synthetic images, search for their closest match in the training set, and plot the duplication histogram for these ``match'' images.   Surprisingly, we see that a typical random image from the dataset is duplicated 11.6 times, which is {\em more} often than a typical matched image, which is duplicated 3.1 times.  However, if we look only at very close matches ($>.5$ SSCD), these match images are replicated on average 34.1 times -- far more often than a typical image.  It seems that replicated content tends to be drawn from training images that are duplicated more than a typical image.


\section{Possible causes of replication}
Data replication in generative models is not inevitable; previous studies of GANs have not found it, and our study of ImageNet LDM did not find any evidence of significant data replication.  What makes {\em Stable Diffusion} different?

We note that both ImageNet LDM (\Cref{sec:imagenet_ldm}) and {\em Stable Diffusion} are built using similar update rules and training routines, and contain similar numbers of parameters.  This rules out a number of factors that may contribute to their difference in behavior.

The most obvious culprit is image duplication within the training set. However this explanation is incomplete and oversimplified; Our models in \Cref{sec:ddpm_training_section} consistently show strong replication when they are trained with small datasets that are unlikely to have any duplicated images.  Furthermore, a dataset in which all images are unique should yield the same model as a dataset in which all images are duplicated 1000 times, provided the same number of training updates are used.


We speculate that replication behavior in {\em Stable Diffusion} arises from a complex interaction of factors, which include that it is text (rather than class) conditioned, it has a highly skewed distribution of image repetitions in the training set, and the number of gradient updates during training is large enough to overfit on a subset of the data.  




\section{Limitations \& Conclusion}
The goal of this study was to evaluate whether diffusion models are capable of reproducing high-fidelity content from their training data, and we find that they are. While typical images from large-scale models do not appear to contain copied content that was detectable using our feature extractors, copies do appear to occur often enough that their presence cannot be safely ignored; {\em Stable Diffusion} images with dataset similarity $\ge .5,$ as depicted in \cref{fig:sd_train_prompts_copies}, account for approximate $1.88\%$ of our random generations. 

Note, however, that our search for replication in {\em Stable Diffusion} only covered the 12M images in the LAION Aesthetics v2 $6+$ dataset. The model was first trained on over $2$ billion images, before being fine-tuned on the 600M LAION Aesthetics V2 $5+$ split.  The dataset that we searched in our study is a small subset of this fine-tuning data, comprising less than $0.6\%$ of the total training data. 


Examples certainly exist of content replication from sources outside the 12M LAION Aesthetics v2 $6+$ split -- see Fig \ref{fig:scream_failure}.  Furthermore, it is highly likely that replication exists that our retrieval method is unable to identify.  For both of these reasons, the results here systematically under-estimate the amount of replication in {\em Stable Diffusion} and other models.

\section{Acknowledgements}
This work was supported by the ONR MURI program, Office of Naval Research (N000142112557), the AFOSR MURI program, DARPA GARD (HR00112020007), and the National Science Foundation (IIS-2212182 \& DMS-1912866). Further support was provided by Capital One Bank.

{\small
\bibliographystyle{ieee_fullname}
\bibliography{egbib}
}
\appendix

\twocolumn[
\centering
\Large
\textbf{Diffusion Art or Digital Forgery? Investigating Data Replication in Diffusion Models} \\
\vspace{0.5em}Supplementary Material \\
\vspace{1.5em}
] 

\section{Additional Stable Diffusion results}

\begin{figure*}[ht]
    \centering
    \includegraphics[width=\linewidth]{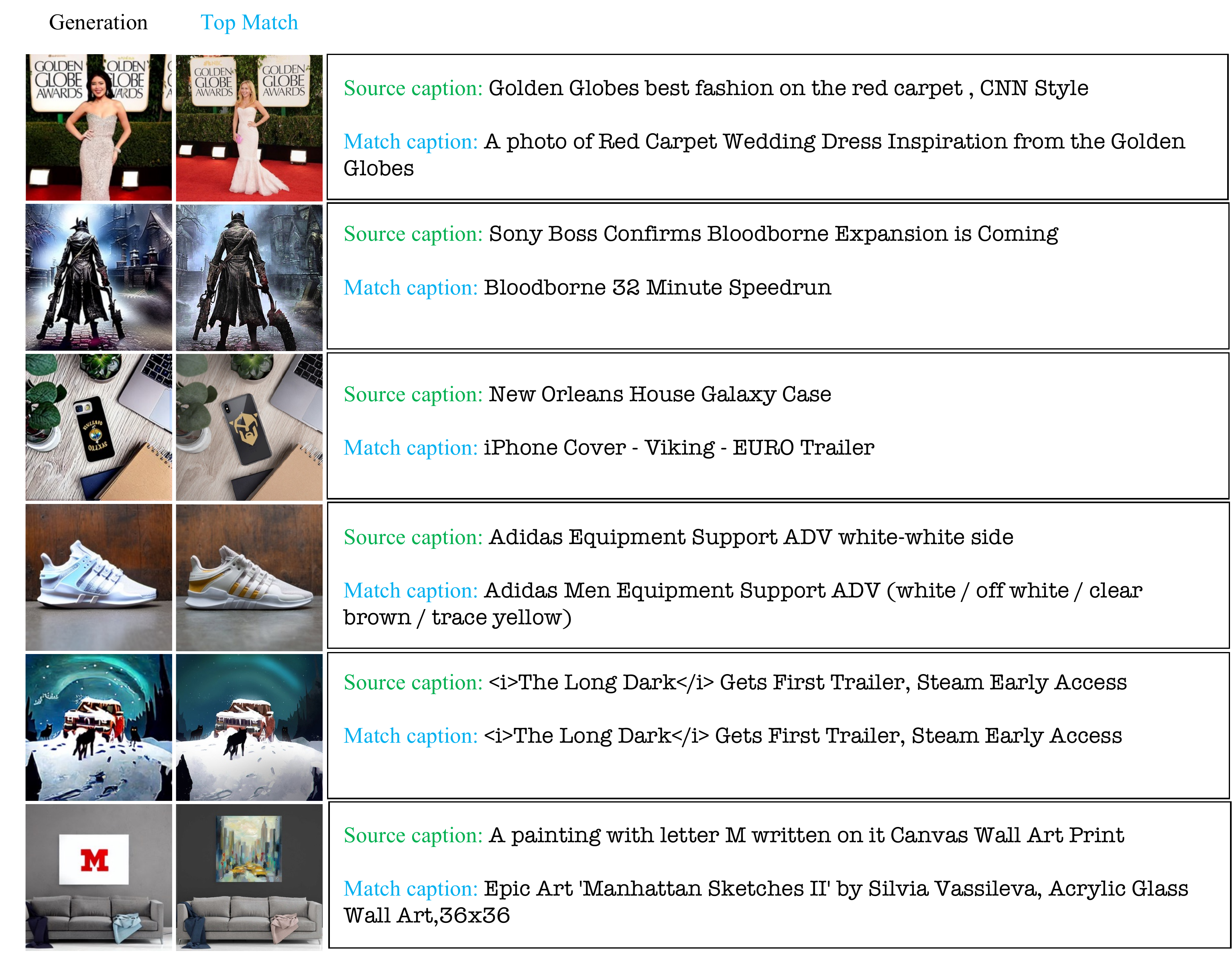}
    \caption{Expanded version of teaser figure (\cref{fig:teaser}). We present the caption used for generation and the caption of the top match.}
    \label{fig:teaser_extend}

\end{figure*}

Here we provide details for the prompts used in the Stable Diffusion section of the main paper. 
We present the prompts used for figures in the main paper in the following tables. We also present additional figures to accompany the observations in the main paper.
\begin{enumerate}
    \item We present the source caption and the match caption for \cref{fig:teaser} in \cref{fig:teaser_extend}. 
    \item We present the prompts used to generate \cref{fig:sd_keyphrase_replic} in \cref{tab:sofa_prompts} and \cref{tab:wave_prompts} in the same order as the figure.

    \item We present the 20 hand-curated prompts for generating famous paintings which we used to understand the copying of content and style in \cref{tab:painting_prompts}. The 6 prompts used to generate \cref{fig:sd_paintings_copies} are on the top part of the table in the same order as they appear in the figure.
\end{enumerate}

\section{Trained diffusion models}
\subsection{Hyperparameters for training}

\paragraph{Celeb-A} We trained the $300$ image model using a batch size of $4$ for $4000$ epochs with a learning rate of $5e-5$ and $200$ warmup steps. For the model with $3000$ images, we used a batch-size of $20$, learning rate of $5e-4$ and $200$ warmup steps, and trained for $4000$ epochs. We used the official pre-trained checkpoint from \url{https://huggingface.co/google/ddpm-celebahq-256} for the whole dataset.

\paragraph{Oxford Flowers} We trained the $100$-image model for $2000$ epochs with a batch size of $5,$ learning rate of $5e-5,$ and $100$ warmup steps. The $1000$-image model is trained with batch size $20$ for $2000$ epochs with learning rate $1e-4$ and with $400$ warmup steps. The model with all data is trained for $1000$ epochs with batch size $32$, learning rate $1e-4$, and $100$ warmup steps.

\section{Stable Diffusion settings}

For the Stable Diffusion experiment, we sampled 9000 captions from whole LAION-Aesthetics (v.6+) dataset with replacement. For each caption, we produced a single generation and did not handpick the generations. The following hyperparameters are used:  \texttt{guidance scale=$7.5$}, \texttt{strength=$0.5$}, \texttt{steps=$50$}.

\begin{table}[!ht]
  \centering
  \caption{Prompts used in generation of \cref{fig:sd_keyphrase_replic} (Top row)}
  \label{tab:sofa_prompts}
  \begin{tabular}{@{}p{0.95\linewidth}@{}}
    \toprule
    Prompts \\
    \midrule
    Florian Pond Behind The House Garden Painting Canvas Wall Art Print  \\
    Starry night painting Canvas Wall Art Print \\
    A poster with letter V Canvas Wall Art Print \\
    A painting of an avocado chair Canvas Wall Art Print  \\
    \bottomrule
  \end{tabular}
\end{table}

\begin{table}[!ht]
  \centering
  \caption{Prompts used in generation of \cref{fig:sd_keyphrase_replic} (Bottom row)}
  \label{tab:wave_prompts}
  \begin{tabular}{@{}p{0.95\linewidth}@{}}
    \toprule
    Prompts \\
    \midrule
    The Great Wave off Kanagawa by Katsushika Hokusai \\
    The Great Wave off Kanagawa \\
    A painting of Kanagawa coast \\
    A painting of a Wave \\
    \bottomrule
  \end{tabular}
  
\end{table}

\begin{table}[]
\centering
  \caption{We show the 20 manually curated prompts we used to understanding the copying of style vs content. The 6 prompts at the top of list are used (in order) to make \cref{fig:sd_paintings_copies}. }
  \label{tab:painting_prompts}
  \begin{tabular}{@{}p{0.95\linewidth}@{}}
    \toprule
    Prompts \\
    \midrule
Starry Night by Vincent van Gogh                       \\
The kiss by Gustav Klimt                                              \\
A Sunday Afternoon on the Island of La Grande Jatte by Georges Seurat \\

Christ And The Storm by Rembrandt                                     \\
Impression, Sunrise by Claude Oscar Monet                \\
Daydream or Reverie by Alphonse Mucha                                 \\
Jean-michel Basquiat by Richard Day                                   \\
\midrule
Bal du moulin de la Galette by Pierre-Auguste Renoir                  \\
Under the Tree of Life by Gustav Klimt                                \\
Horn of Babel by Vladimir Kush                                        \\
Man in a Bowler Hat  by Rene Magritte                                 \\
The Autumn by Alphonse Mucha                                          \\

Blue Lovers by Marc Chagall                                           \\
Marilyn Monroe By Andy Warhol                                         \\
Dora Maar au Chat By Pablo Picasso                                    \\

A girl with pearl earring by Johannes Vermeer                         \\
The Persistence of Memory by Salvador Dali                            \\
Ophelia by Sir John Everett Millais                                   \\

Cafe Terrace at Night by Vincent van Gogh                             \\

The Scream by Edvard Munch                                          \\ 
    \bottomrule
  \end{tabular}
\end{table}

\begin{figure}[h]
    \centering
    \includegraphics[width=\columnwidth]{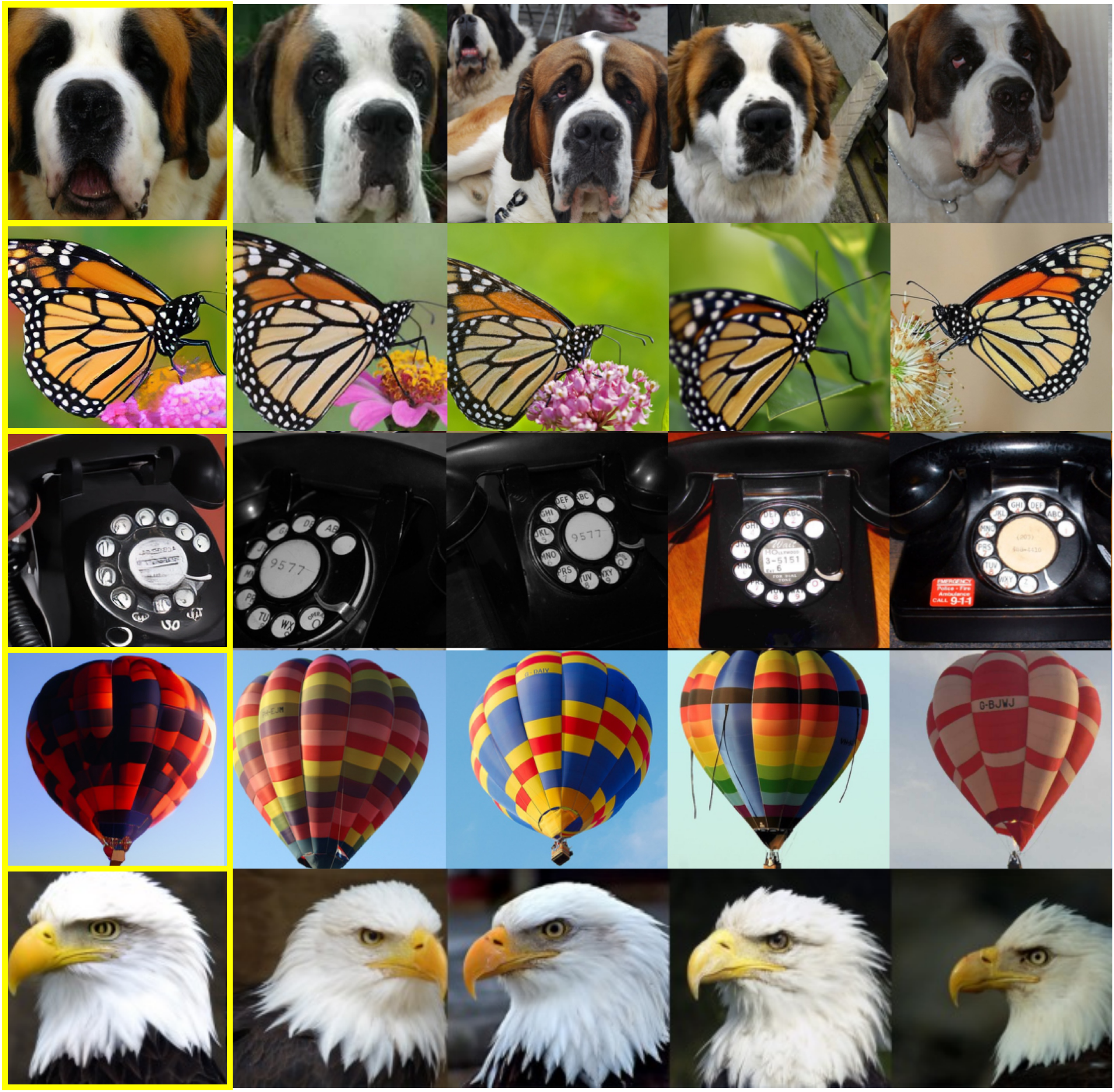}
    \caption{We present a few generations (in yellow boxes) from the ImageNet conditional LDM model and the top 4 matches from the ImageNet training set using DINO split-product.}
    \label{fig:imagenet_matches}
\end{figure}

\subsection{Misc results}

\begin{figure*}[ht]
    \centering
    \includegraphics[width=\linewidth]{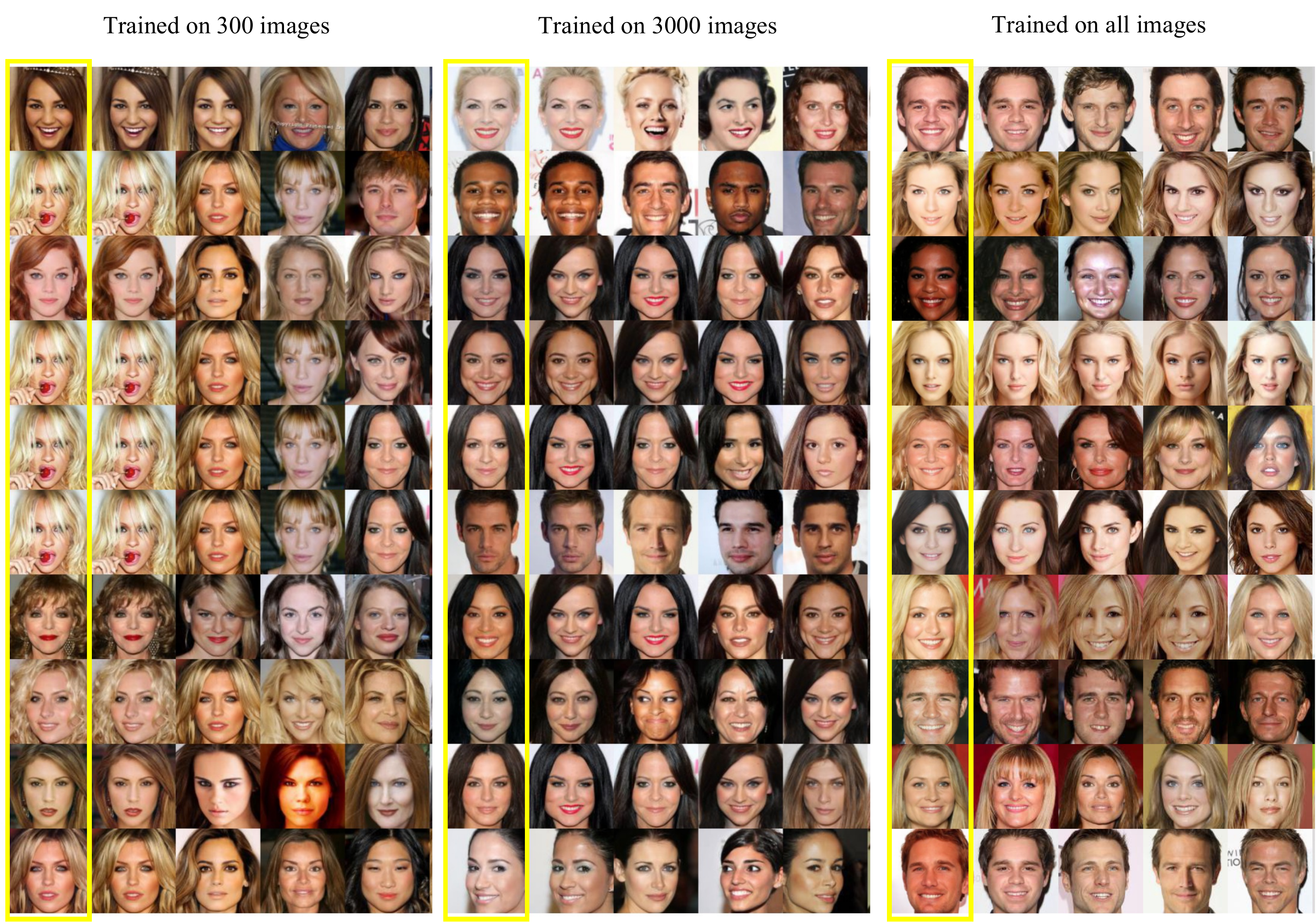}
    \caption{We show the expanded version of \cref{fig:celeba_matches} for DINO split-product. The generations are in yellow bounding boxes and we show the top 4 matches per image.}
    \label{fig:celeba_dino_expanded}
\end{figure*}

We also include the top 10 generations and their top 5 matches for DINO split-product on the Celeb-A dataset for all 3 diffusion models (trained on 300, 3000 and 30k images) in \cref{fig:celeba_dino_expanded}. We further present a few generations from the pretrained ImageNet conditional-LDM model and their closest matches in the ImageNet train set in \cref{fig:imagenet_matches}.

\end{document}